\documentclass[letterpaper]{article} 
\usepackage{aaai25}  
\usepackage{times}  
\usepackage{helvet}  
\usepackage{courier}  
\usepackage[hyphens]{url}  
\usepackage{graphicx} 
\usepackage{natbib}  
\usepackage{caption} 
\usepackage{algorithm}
\usepackage{algpseudocode}

\usepackage{multirow,makecell,booktabs}
\usepackage{color}
\usepackage{url}
\usepackage{subfig}
\usepackage[T1]{fontenc}
\usepackage{comment}
\usepackage{graphicx}
\usepackage[export]{adjustbox}
\usepackage{float,dblfloatfix}
\usepackage{rotating,changepage}
\usepackage{arydshln}

\usepackage{soul}
\let\ul\underline

\usepackage{enumitem}

\usepackage{array}
\newcolumntype{L}[1]{>{\raggedright\let\newline\\\arraybackslash\hspace{0pt}}m{#1}}
\newcolumntype{C}[1]{>{\centering\let\newline\\\arraybackslash\hspace{0pt}}m{#1}}
\newcolumntype{R}[1]{>{\raggedleft\let\newline\\\arraybackslash\hspace{0pt}}m{#1}}

\usepackage{amsmath,amsfonts,bm}

\def\eqref#1{eq.~(\ref{#1})}
\def\Eqref#1{Eq.~(\ref{#1})}

\def\1{\bm{1}}

\def\rd{{\textnormal{d}}}

\def\vx{{\bm{x}}}

\def\vz{{\bm{z}}}

\DeclareMathAlphabet{\mathsfit}{\encodingdefault}{\sfdefault}{m}{sl}
\SetMathAlphabet{\mathsfit}{bold}{\encodingdefault}{\sfdefault}{bx}{n}

\def\gF{{\mathcal{F}}}
\def\gG{{\mathcal{G}}}

\usepackage{tikz}
\usetikzlibrary{shapes,arrows,positioning,fit,backgrounds}

\usepackage{newfloat}
\usepackage{listings}
\DeclareCaptionStyle{ruled}{labelfont=normalfont,labelsep=colon,strut=off} 
\floatstyle{ruled}
\newfloat{listing}{tb}{lst}{}
\floatname{listing}{Listing}
\title{\texttt{DualDynamics}: Synergizing Implicit and Explicit Methods for Robust Irregular Time Series Analysis}
\author{
    YongKyung Oh\textsuperscript{\rm 1}\thanks{These two authors are equal contributors to this work and designated as co-first authors.}, Dong-Young Lim\textsuperscript{\rm 2,3}\footnotemark[1], Sungil Kim\textsuperscript{\rm 2,3}\thanks{Corresponding Author}
}

\affiliations{
    \textsuperscript{\rm 1}Medical \& Imaging Informatics (MII) Group, 
    University of California, Los Angeles (UCLA), CA, USA \\
    \textsuperscript{\rm 2}Department of Industrial Engineering, Ulsan National Institute of Science and Technology (UNIST), Republic of Korea \\
    \textsuperscript{\rm 3}Artificial Intelligence Graduate School, Ulsan National Institute of Science and Technology (UNIST), Republic of Korea \\
    yongkyungoh@mednet.ucla.edu, \{dlim, sungil.kim\}@unist.ac.kr
}

\begin{document}

\maketitle

\begin{abstract}
    Real-world time series analysis faces significant challenges when dealing with irregular and incomplete data. While Neural Differential Equation (NDE) based methods have shown promise, they struggle with limited expressiveness, scalability issues, and stability concerns. Conversely, Neural Flows offer stability but falter with irregular data. We introduce \texttt{DualDynamics}, a novel framework that synergistically combines NDE-based method and Neural Flow-based method. This approach enhances expressive power while balancing computational demands, addressing critical limitations of existing techniques. We demonstrate \texttt{DualDynamics}' effectiveness across diverse tasks: classification of  robustness to dataset shift, irregularly-sampled series analysis, interpolation of missing data, and forecasting with partial observations. Our results show consistent outperformance over state-of-the-art methods, indicating \texttt{DualDynamics}' potential to advance irregular time series analysis significantly. 
\end{abstract}

%
\begin{links}
    \link{Code}{https://github.com/yongkyung-oh/DualDynamics}
    \link{Extended version}{https://arxiv.org/abs/2401.04979}
\end{links}

\section{Introduction}

Effectively modeling time series data is a cornerstone in machine learning, supporting numerous applications across diverse sectors. This significance has driven substantial research endeavors aimed at developing novel methodologies capable of accurately capturing the complexities of continuous latent processes~\citep{chen2018neural,kidger2020neural,kidger2022neural,oh2024stable}.

\textcolor{black}{Recent advancements have led to two distinct approaches: \emph{implicit} methods, such as Neural Differential Equation (NDE)-based techniques, and \emph{explicit} methods, such as Neural Flows.}
Neural Ordinary Differential Equations (Neural ODEs)~\citep{chen2018neural}, Neural Controlled Differential Equations (Neural CDEs)~\citep{kidger2020neural} and Neural Stochastic Differential Equations (Neural SDEs)~\citep{oh2024stable} have emerged as prominent implicit methods, recognized for their ability to learn continuous-time dynamics and underlying temporal structures. These approaches directly learn continuous latent representations based on vector fields parameterized by neural networks.
However, NDE-based methods face challenges including limited expressive power~\citep{dupont2019augmented, massaroli2020dissecting, chalvidal2021go,kidger2022neural} and scalability issues when analyzing irregular or complex time series data~\citep{norcliffe2020neural,morrill2021neural,morrill2022choice,irie2022neural, pal2023controlled}. These limitations are attributed to data complexity, sequence length variations, and the stability constraints of numerical solvers~\citep{he2023incremental, westny2023stability}.

In parallel, explicit methods have been developed, focusing on directly mapping solution curves using neural networks~\citep{lu2018beyond,sonoda2019transport,massaroli2020stable,bilovs2021neural}. These approaches offer invertible solutions and enhanced stability through the change of variables formula~\citep{kobyzev2020normalizing,papamakarios2021normalizing}. However, they struggle with irregularly-sampled time series and are sensitive to initial state and corresponding initial value problem. 

To address the limitations of both implicit and explicit methods, we introduce \texttt{DualDynamics}, a novel framework that synergistically combines NDE-based model and Neural Flow-based model. This approach aims to leverage the strengths of both paradigms: the flexibility of implicit methods in handling irregular data and the stability and computational efficiency of explicit methods. By integrating these complementary approaches, \texttt{DualDynamics} seeks to enhance expressive power, improve scalability, and maintain stability in modeling complex, irregular time series data.

\section{Related Works}\label{Related_work}
The landscape of neural differential equations for time series modeling has been dominated by two distinct approaches: implicit methods, exemplified by Neural ODEs, Neural CDEs and Neural SDEs, and explicit methods, such as Neural Flows. These approaches fundamentally differ in how they represent and solve the underlying \textcolor{black}{temporal dynamics}. 

\subsection{Implicit Methods for Continuous Dynamics}
Implicit methods in deep learning for time series modeling solve differential equations numerically, allowing for adaptive computation and flexible handling of irregular data. These methods have gained significant attention due to their ability to model complex dynamics.

Let $\vx = (x_0, x_1, \ldots, x_n)$ be a vector of original irregularly-sampled observations and $\vz(t)$ denote a latent state at time $t$. Continuous latent dynamics $\vz(t)$ can be achieved by integration of $\rd \vz(t)$ over time in $[0,t]$.
Neural Differential Equations can be expressed as:
\begin{align}
\text{Neural ODEs: } \quad
\rd \vz(t) &= f(s,\vz(s);\theta_f) \, \rd s \label{eq:neural_ode} \\
\text{Neural CDEs: } \quad
\rd \vz(t) &= f(s,\vz(s);\theta_f) \, \rd X(s) \label{eq:neural_cde} \\
\text{Neural SDEs: } \quad
\rd \vz(t) &= f(s,\vz(s);\theta_f) \, \rd s \nonumber \\ &+ g(s,\vz(s);\theta_g) \, \rd W(s) \label{eq:neural_sde}
\end{align}
with initial condition $\vz(0) = h(\vx;\theta_{h})$, where $h: \mathbb{R}^{d_x} \rightarrow \mathbb{R}^{d_z}$ is an affine function. $f$ and $g$ are neural networks parameterized by $\theta_f$ and $\theta_g$, respectively. $X(t)$ is a control path for Neural CDEs, and $W(t)$ is a Wiener process (or Brownian motion) for Neural SDEs.

Neural ODEs~\citep{chen2018neural} model deterministic dynamics but may struggle with expressiveness and stability for complex trajectories~\citep{dupont2019augmented, massaroli2020dissecting}. 
Furthermore, Neural CDEs~\citep{kidger2020neural} handle irregular time series effectively but can be computationally intensive for long sequences~\citep{morrill2021neural}. Therefore, recent works introduced extended architecture using Neural CDE.  
On the other hand, Neural SDEs~\citep{li2020scalable} incorporate uncertainty modeling but face challenges in training and inference due to their stochastic nature~\citep{jia2019neural, oh2024stable}.
These methods offer varying approaches to modeling continuous-time dynamics in time series data, each with its own strengths and limitations in handling complexity, irregularity, and uncertainty.

\subsection{Explicit Methods for Continuous Dynamics}
Explicit methods directly model the solution of differential equations, often providing faster computation and improved stability, albeit potentially at the cost of some flexibility.
Neural Flows, introduced by \citet{bilovs2021neural},  present a novel approach to directly model the trajectory of the latent state $\vz(t)$, in contrast to Neural ODEs which represent the rate of change of $\vz(t)$ over time in an integration form. Specifically, $\vz(t)$ is expressed as:
\begin{align}\label{eq:neural_flow}
\vz(t) & = \gF(t, \vz(0); \theta_\gF),  
\end{align}
with the initial value $\vz(0) = h(\vx;\theta_{h})$, where $\gF$ is a neural network parameterized by $\theta_{\mathcal F}$. Note that while $\mathcal F$ represents the solution to the initial value problem of Neural ODEs, as described in \Eqref{eq:neural_ode}, the formulation of Neural Flows in \Eqref{eq:neural_flow} does not require the use of an ODE solver. 
However, the requirement for the Neural Flow $\gF(\cdot, \vz(0))$ to be invertible restricts the type of neural network architectures, which may not be suitable if the initial value is unknown or unstable to estimate.
In practice, determining the exact flow is challenging since many real-world problems do not have analytical solutions, which can lead to training instability.

\section{Methodology}\label{Methodology}

\subsection{Proposed Framework}
To illustrate our \texttt{DualDynamics} framework, we utilize Neural CDEs as the primary example in this section. Neural CDEs serve as an effective representation of the implicit component in our dual approach, offering a robust foundation for modeling irregular time series data. 

\begin{figure}[htbp]
\centering\captionsetup{skip=5pt}
\captionsetup[subfloat]{skip=5pt}
  \includegraphics[width=\linewidth]{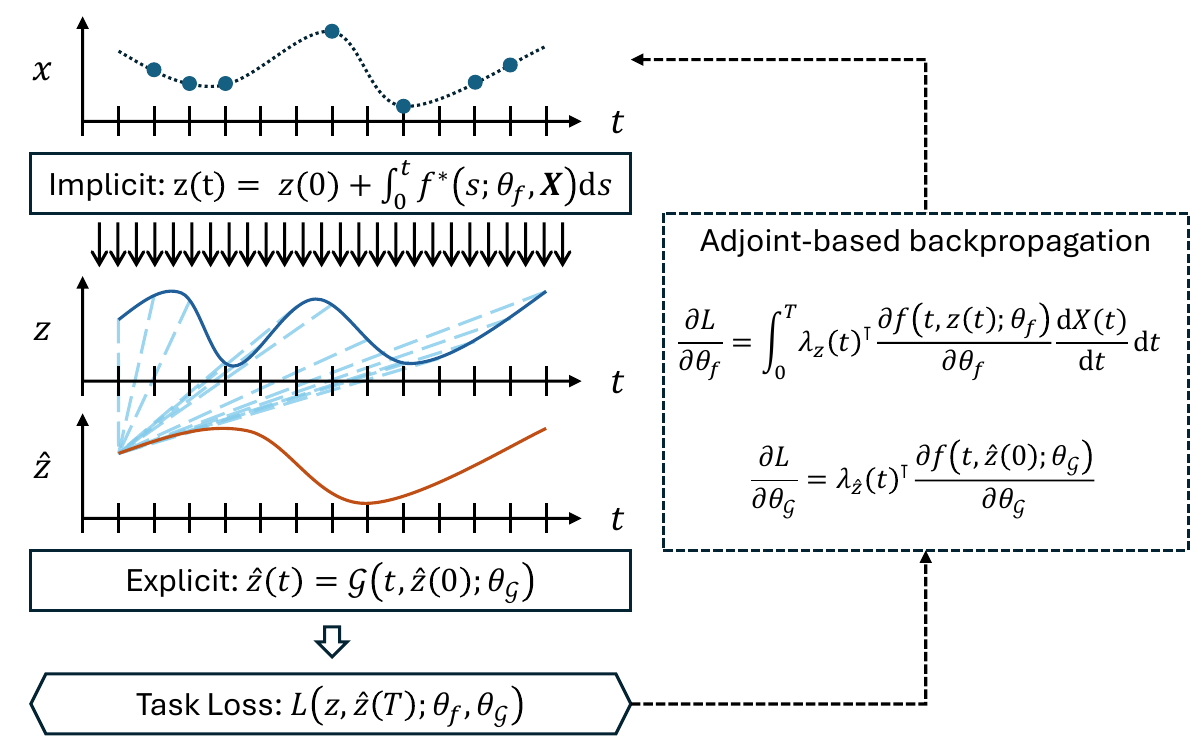}
\caption{Overview of the proposed \texttt{DualDynamics}}\label{fig:overview}
\end{figure}

Figure~\ref{fig:overview} shows the conceptual overview of the proposed architecture using Neural CDE (implicit) and Flow-based model (explicit) for classification task.
 
\textcolor{black}{
The proposed method combines the stability of flow models with NDE-based model to achieve a better representation of the irregular time series. The flow model, known for its capability to smoothly transform complex distributions, offers inherent stability when dealing with time-varying dynamics.
To achieve stability in the latent representation $\vz(t)$, we introduce a secondary latent space, denoted by $\hat{\vz}(t)$. This secondary latent space is inspired by the principles of Neural Flow and is designed to produce a regularized representation of the original latent variable.}
From \Eqref{eq:neural_cde}, the latent representation $\vz(t)$ can be derived from an initial vector $\vz(0)$, which is transformed from the raw input $x_0$. %
Thus, the proposed dual latent variable evolves in time as:
\begin{align}\label{eq:cde_part}
\vz(t) & = \vz(0) + \int_0^t f(s,\vz(s); \theta_f) \, \rd X(s), 
\end{align}
where the initial value of the first latent representation $\vz(0) = h(x_0;\theta_{h})$, and $h: \mathbb{R}^{d_x} \rightarrow \mathbb{R}^{d_z}$.
As stated in \citet{chen2018neural} and \citet{kidger2020neural}, we focus on the time-evolving state $\vz$ from the linear mapping from $\vx$ using $h$. \Eqref{eq:cde_part} can be rewritten as follows:
\begin{align}\label{eq:cde_part2}
    \vz(t) & = \vz(0) + \int_0^t f^{*}(s; \theta_{f}, X) \, \rd s,
\end{align}
where $f^{*}(s;\theta_f, X):=f(s,\vz(s); \theta_f) \frac{\rd X(s)}{\rd s}$.  Recall that $X$ is generated from $\vx$.  Here, we implement $X$ for `Hermite Cubic splines with a backward difference', which is smooth and unique, as discussed by \citet{morrill2022choice}. Through this part, we find $\vz \in \mathbb{R}^{d_z}$, and then plug it into the following flow model $\gG$ as:
\begin{align}\label{eq:flow_part}
\hat{\vz}(t) & = \gG(t, \hat{\vz}(0); \theta_\gG),
\end{align}
with $\hat{\vz}(0) = k(\vz;\theta_{k})$, and $k: \mathbb{R}^{d_z \times T} \rightarrow \mathbb{R}^{d_z}$. In other words, $\hat{\vz}(0)$ is determined by $\vz(t)$ for $t\in[0,T]$.

Here, $\gG$ satisfies the ODE defined in \Eqref{eq:cde_part2}, i.e., $\frac{\mathrm{d} \gG(t,\hat{\vz}(0))}{\mathrm{d} t} =f^*(t;\theta_f,X)$ and $\gG(\cdot, \hat{\vz}(0))$ is invertible. 
In summary, after obtaining the primary latent space $\vz(t)$, we compute $\gG$ with the initial value $\hat{\vz}(0)$. This model generates the secondary latent state $\hat{\vz}(t)$ without the need to directly determine its derivative $\frac{\rd \hat{\vz}(t)}{\rd t}$.

\paragraph{Expressive Power of Flow Model $\gG$}

Flow-based model, which is invertible and differentiable (known as `diffeomorphisms'), has the capability of representing any distribution. Given the Neural Flow $\gG$, there exists a unique input $\hat{\vz}(0)$ such that, $\hat{\vz}(0) = \gG^{-1}(t, \hat{\vz}(t); \theta_\gG)$. This invertibility ensures a one-to-one mapping between the initial and transformed states, preserving information content.
The change of variables formula for probability densities leads to:
\begin{align}\label{eq:det}
    p_{\hat{\vz}(t)}(\hat{\vz}(t)) = p_{\hat{\vz}(0)}(\hat{\vz}(0)) \left|\det \bm{J}_{\gG}(\hat{\vz}(0))\right|^{-1},    
\end{align}
where $\bm{J}_{\gG}(\hat{\vz}(0))$ is the Jacobian matrix of $\gG$ at $\hat{\vz}(0)$, and $\left|\det \bm{J}_{\gG}(\hat{\vz}(0))\right|$ is its determinant. 

Suppose $\hat{\vz}(0) \sim \mathcal{U}([0,1]^{d_z})$, meaning each component of $\hat{\vz}(0)$ is uniformly distributed between 0 and 1. The probability density function of $\hat{\vz}(0)$ in this multi-dimensional space is constant, $p_{\hat{\vz}(0)}(y) = 1$ if $y \in [0,1]^{d_z}$.
Then, \Eqref{eq:det} can be simplified to:
\begin{align}
p_{\hat{\vz}(t)}(\hat{\vz}(t)) = \left| \det \bm{J}_{\gG}(\hat{\vz}(0)) \right|^{-1}.
\end{align}
The determinant of the Jacobian $\left| \det \bm{J}_{\gG}(\hat{\vz}(0)) \right|$ accounts for the `stretching' or `squeezing' of the volume elements during the transformation. By adjusting $\theta_\gG$, the model can control how much each part of the space $[0,1]^{d_z}$ expands, contracts or projects, allowing $\gG$ to mold the uniform distribution into any desired complex distribution $p_{\hat{\vz}(t)}$.

\subsection{Properties of the Proposed Method}\label{Properties}

\paragraph{Preservation of Probability Density}
\textcolor{black}{It is crucial for ensuring that the total probability across transformations remains constant, enabling accurate modeling of dynamic systems' evolution without loss or gain of information~\citep{li2008principle,dinh2016density}. }
For each fixed $t$, $p_{\vz(t)}$ and $p_{\hat{\vz}(t)}$ be the probability density functions of $\vz(t)$ and $\hat{\vz}(t)$, respectively. From the \emph{instantaneous change of variables formula}~\citep{chen2018neural}, we observe that: 
\begin{align}
    \frac{\partial \log p_{\vz(t)}(\vz(t)) }{\partial t} &= -\text{Tr}\left(\bm{J}_{f^{*}}(\vz(t))\right), \label{eq:density1}\\
    \frac{\partial \log p_{\hat{\vz}(t)}(\hat{\vz}(t)) }{\partial t} &= -\text{Tr}\left(\bm{J}_{\frac{\mathrm{d}\gG}{\mathrm{d} t}}(\hat{\vz}(t))\right),    \label{eq:density2}
\end{align}
where $Tr(A)$ is the trace of matrix $A$ and $\bm{J}_{f^*}$ is the Jacobian matrix of $f^*$. 
$\bm{J}_{\frac{\mathrm{d}\gG}{\mathrm{d} t}}$ is the Jacobian matrix of $\frac{\mathrm{d}\gG}{\mathrm{d} t}$ and we have used $\frac{\mathrm{d} \gG(t,\hat{\vz}(0))}{\mathrm{d} t} =f^*(t;\theta_f,X)$ for all $t$.
Therefore, the total changes in log-density for $\vz$ and $\hat{\vz}$ from time $0$ to $t$ are given by:
\begin{align}
    &\log p_{\vz(t)}(\vz(t)) - \log p_{\vz(0)}(\vz(0)) \nonumber \\
    &\qquad =- \int_{0}^{t} \text{Tr}\left(\bm{J}_{f^{*}}(\vz(s))\right) \, \rd s,\label{eq:density_total1}\\   
    &\log p_{\hat{\vz}(t)}(\hat{\vz}(t)) - \log p_{\hat{\vz}(0)}(\hat{\vz}(0)) \nonumber \\
    &\qquad =- \int_{0}^{t} \text{Tr}\left(\bm{J}_{\frac{\mathrm{d}\gG}{\mathrm{d} s}}(\hat{\vz}(s))\right) \, \rd s, \label{eq:density_total2-1} \\
    &\qquad =- \int_{0}^{t} \text{Tr}\left(\bm{J}_{f^*}(\hat{\vz}(s))\right) \, \rd s,\label{eq:density_total2}
\end{align}
\Eqref{eq:density_total1} and \Eqref{eq:density_total2} imply that the probability content of corresponding regions in the primary latent space $\vz$ and the second latent space $\hat \vz$ must be the same when the Neural Flow $\gG$ is perfectly estimated~\citep{kingma2018glow,onken2021ot}.  

\paragraph{Computational Advantage of Flow Model $\gG$}
We are interested in computing the trace of $\bm{J}_{\frac{\mathrm{d}\gG}{\mathrm{d} s}}$ in \Eqref{eq:density_total2-1}, which is a key quantity to understand the change in log-density. 
We use Hutchinson's trace estimator\footnote{
Hutchinson's trace estimator states that for a square matrix $A$, the trace of $A$ can be estimated as: $\text{Tr}(A) \approx \mathbb{E}_{\bm{v}}[\bm{v}^\top A \bm{v}],$ where $\bm{v}$ is a random vector with each element drawn independently from a distribution with zero mean and unit variance.}~\citep{hutchinson1989stochastic} used to estimate the trace of a matrix efficiently, especially in situations where the matrix is large and computing the trace directly is computationally expensive\footnote{Hutchinson's trace estimator significantly reduces computational complexity from $\mathcal{O}({d_z}^2)$ for direct trace computation to $\mathcal{O}(d_z)$ for matrix-vector multiplications.}~\citep{adams2018estimating,han2015large,grathwohl2018ffjord}.

Applying Hutchinson's trace estimator to $\bm{J}_{\frac{\mathrm{d}\gG}{\mathrm{d} s}}$, we have:
\begin{align}    \text{Tr}\left(\bm{J}_{\frac{\mathrm{d}\gG}{\mathrm{d} t}}(\hat{\vz}(t))\right) \approx \mathbb{E}_{\bm{v}}\left[\bm{v}^\top \bm{J}_{\frac{\mathrm{d}\gG}{\mathrm{d} t}}(\hat{\vz}(t)) \bm{v}\right].
\end{align}
Then, \Eqref{eq:density_total2-1} can be rewritten as follows:
\begin{align}
    &\log p_{\hat{\vz}(t)}(\hat{\vz}(t)) - \log p_{\hat{\vz}(0)}(\hat{\vz}(0)) \nonumber \\
    &\qquad \approx -\mathbb{E}_{\bm{v}}\left[ \int_{0}^{t} \bm{v}^\top \bm{J}_{\frac{\mathrm{d}\gG}{\mathrm{d} t}}(\hat{\vz}(t))\bm{v}  \, \rd s \right] \nonumber \\
    &\qquad = -\mathbb{E}_{\bm{v}}\left[ \int_{0}^{t} \bm{v}^\top \frac{\partial \bm{J}_{\gG}(\hat{\vz}(t))}{\partial t} \bm{v}  \, \rd s \right]. 
\end{align}
where the last inequality is obtained by changing the order of partial derivatives. 
\textcolor{black}{Incorporating the flow model $\gG$ into the original NDE-based model enhances its expressive power without significantly increasing the computational burden, as opposed to the addition of an extra NDE formulation.}

\subsection{Parameter Optimization}\label{Optimization}
To optimize parameters involved in the proposed method, we employ adjoint-based backpropagation~\citep{chen2018neural,kidger2020neural,kidger2022neural,oh2024stable}. This method allows for efficient gradient computation, ensuring that the information from one step influences and refines the subsequent step, leading to a more unified optimization process. 
The adjoint method is a powerful technique often used to calculate the gradients of systems governed by differential equations. The basic idea is to recast the derivative computation problem as the adjoint equation \cite{pontryagin2018mathematical,pollini2018adjoint}.

Consider an objective function $\mathcal{L}(\vz, \hat{\vz}(T); \theta_f, \theta_{\gG})$ that depends on the final value of the second latent state, $\hat \vz(T)$. Then, one can define the adjoint state $\bm{\lambda}_{\vz}(t)$ and $\bm{\lambda}_{\hat{\vz}}(t)$ as the gradient of the loss function with respect to the state $\vz(t)$ and $\hat{\vz}(t)$:
\begin{align}
    \bm{\lambda}_{\vz}(t) = \frac{\partial \mathcal{L}}{\partial \vz(t)}, \qquad \bm{\lambda}_{\hat{\vz}}(t) = \frac{\partial \mathcal{L}}{\partial \hat{\vz}(t)}.
\end{align}
To estimate the model parameters $\theta_f$ and $\theta_\gG$, we compute the gradients of the loss function with respect to these parameters simultaneously:
\begin{align}
   \frac{\partial \mathcal{L}}{\partial \theta_f} &= \int_0^T \bm{\lambda}_{\vz}(t)^\top \frac{\partial f(t, \vz(t); \theta_f)}{\partial \theta_f} \, \frac{\rd X(t)}{\rd t} \rd t,\\ 
   \frac{\partial \mathcal{L}}{\partial \theta_\gG} &= \bm{\lambda}_{\hat{\vz}}(T)^\top \frac{\partial \gG(T, \hat{\vz}(0); \theta_{\gG})}{\partial \theta_\gG}.
\end{align}
Note that $\bm{\lambda}_{\vz}(t)$ can be derived by integration of adjoint dynamics backward in time from $T$ to 0, which is the continuous-time analog to the backpropagation algorithm~\citep{rumelhart1986learning,kidger2021hey}. On the other hand, $\bm{\lambda}_{\hat{\vz}}(t)$ is derived by backpropagation algorithm directly. 

The \texttt{DualDynamics} framework integrates implicit and explicit components into a unified architecture, enabling concurrent optimization of both elements. This joint learning process ensures that the implicit Neural Differential Equation and the explicit Neural Flow components evolve synergistically, rather than sequentially, thereby capitalizing on their complementary strengths in a single, cohesive optimization procedure.

\section{Experiments}\label{Experiment}

All experiments were performed using a server on Ubuntu 22.04 LTS, equipped with an Intel(R) Xeon(R) Gold 6242 CPU and a cluster of NVIDIA A100 40GB GPUs. The source code for our experiments and datasets can be accessed at \url{https://github.com/yongkyung-oh/DualDynamics}.
\textcolor{black}{Please refer the supplementary material for the experiment settings and implementation details.}

\subsection{Proposed Method}
The implicit component in our framework primarily utilizes Neural CDE, with Neural ODE and Neural SDE explored as alternatives in our ablation study. For the explicit component, we evaluated three distinct flow models: ResNet Flow, GRU Flow, and Coupling Flow. These flow architectures, as outlined by \citet{bilovs2021neural}, were incorporated into our proposed framework. To assess the impact of using flow models, we also implemented a conventional multilayer perceptron (MLP) as a baseline for comparison. In practice, optimal flow model is selected by hyperparameter tuning stage. 

\subsection{Benchmark Methods}
In experiments, we employed a variety of benchmark models for classification and forecasting, incorporating conventional RNN~\citep{rumelhart1986learning,medsker1999recurrent}, LSTM~\citep{hochreiter1997long}, and GRU~\citep{chung2014empirical}, alongside their notable variants, including GRU-$\Delta t$~\citep{choi2016doctor}, and GRU-D~\citep{che2016recurrent}. 
Also, we used attention-based method including transformer~\citep{vaswani2017attention}, 
Multi-Time Attention Networks (MTAN)~\citep{shukla2021multi}, and Multi-Integration Attention Module (MIAM)~\citep{lee2022multi}. 
ODE-based models like, GRU-ODE~\citep{de2019gru}, ODE-RNN~\citep{rubanova2019latent} ODE-LSTM~\citep{lechner2020learning}, Latent-ODE~\citep{rubanova2019latent}, Augmented-ODE~\citep{dupont2019augmented}, and Attentive co-evolving neural ordinary differential equations (ACE-NODE)~\citep{jhin2021ace} are considerd. Furthermore, Neural CDE~\citep{kidger2020neural}, and Neural Rough Differential Equation (Neural RDE)~\citep{morrill2021neural}, were included in our comparative study. 
Additionally, recent advances in Neural CDE are considered, including Attentive Neural Controlled Differential Equation (ANCDE)~\citep{jhin2024attentive}, EXtrapolation and InTerpolation-based model (EXIT)~\citep{jhin2022exit}, and LEArnable Path-based model (LEAP)~\citep{jhin2023learnable}. 
Lastly, variations of Neural Flow~\citep{bilovs2021neural} are also included. 

In the interpolation task, encoder-decoder architecture is implemented by Variational Auto-Encoder (VAE) scheme with the evidence lower bound (ELBO). RNN-VAE~\citep{chen2018neural}, L-ODE-RNN~\citep{chen2018neural}, L-ODE-ODE~\citep{rubanova2019latent}, and mTAND-Full (MTAN encoder--MTAN decoder model)~\citep{shukla2021multi}, were used as benchmark methods, based on the suggestion of \citet{shukla2021multi}. 
In our model, the proposed method is utilized for the encoder component, while the conventional RNN is employed as the decoder.

\begin{table*}[htbp]
\scriptsize\centering\captionsetup{skip=5pt}
\begin{tabular}{@{}lcccccccccc@{}}
\toprule
\multirow{2.5}{*}{\textbf{Methods}} & \multicolumn{2}{c}{\textbf{Regular}}   & \multicolumn{2}{c}{\textbf{30\% Missing}} & \multicolumn{2}{c}{\textbf{50\% Missing}} & \multicolumn{2}{c}{\textbf{70\% Missing}} & \multicolumn{2}{c}{\textbf{Average}}   \\ \cmidrule(lr){2-3} \cmidrule(lr){4-5} \cmidrule(lr){6-7} \cmidrule(lr){8-9} \cmidrule(lr){10-11} 
                                  & \textbf{Accuracy}      & \textbf{Rank} & \textbf{Accuracy}        & \textbf{Rank}  & \textbf{Accuracy}        & \textbf{Rank}  & \textbf{Accuracy}        & \textbf{Rank}  & \textbf{Accuracy}      & \textbf{Rank} \\ \midrule
RNN                               & 0.560 (0.072)          & 11.9          & 0.484 (0.075)            & 14.7           & 0.471 (0.082)            & 14.0           & 0.453 (0.068)            & 14.4           & 0.492 (0.074)          & 13.8          \\
LSTM                              & 0.588 (0.067)          & 11.0          & 0.552 (0.075)            & 10.2           & 0.516 (0.073)            & 11.4           & 0.505 (0.067)            & 11.3           & 0.540 (0.071)          & 11.0          \\
GRU                               & 0.674 (0.080)          & 7.3           & 0.639 (0.065)            & 8.4            & 0.611 (0.076)            & 9.0            & 0.606 (0.088)            & 8.6            & 0.633 (0.077)          & 8.3           \\
GRU-$\Delta t$                            & 0.629 (0.065)          & 9.9           & 0.636 (0.069)            & 8.2            & 0.651 (0.068)            & 7.2            & 0.649 (0.074)            & 7.9            & 0.641 (0.069)          & 8.3           \\
GRU-D                             & 0.593 (0.088)          & 10.9          & 0.579 (0.087)            & 10.6           & 0.580 (0.075)            & 10.6           & 0.599 (0.062)            & 10.1           & 0.588 (0.078)          & 10.6          \\ \midrule
Transformer                       & \ul{0.720 (0.063)}    & \ul{6.8}     & 0.663 (0.066)            & 8.0            & \ul{0.669 (0.075)}      & 7.4            & 0.643 (0.079)            & 8.2            & \ul{0.674 (0.071)}    & 7.6           \\
MTAN                              & 0.654 (0.088)          & 9.4           & 0.634 (0.098)            & 7.4            & 0.631 (0.089)            & \ul{6.8}      & 0.642 (0.073)            & 6.8            & 0.640 (0.087)          & 7.6           \\
MIAM                              & 0.573 (0.079)          & 11.9          & 0.585 (0.086)            & 10.0           & 0.572 (0.071)            & 9.9            & 0.544 (0.063)            & 11.1           & 0.568 (0.075)          & 10.7          \\ \midrule
GRU-ODE                           & 0.663 (0.072)          & 7.6           & 0.661 (0.069)            & 7.4            & 0.664 (0.069)            & 6.8            & \ul{0.659 (0.081)}      & 6.9            & 0.662 (0.073)          & \ul{7.2}     \\
ODE-RNN                           & 0.652 (0.085)          & 7.3           & 0.632 (0.076)            & 7.8            & 0.626 (0.086)            & 7.8            & 0.653 (0.059)            & \ul{6.5}      & 0.641 (0.076)          & 7.4           \\
ODE-LSTM                          & 0.566 (0.074)          & 11.4          & 0.518 (0.069)            & 12.7           & 0.501 (0.068)            & 13.3           & 0.474 (0.068)            & 13.4           & 0.515 (0.070)          & 12.7          \\ \midrule
Neural CDE                        & 0.681 (0.073)          & 7.6           & \ul{0.672 (0.068)}      & 7.4            & 0.661 (0.070)            & 7.3            & 0.652 (0.091)            & 7.3            & 0.667 (0.075)          & 7.4           \\
Neural RDE                        & 0.649 (0.082)          & 8.6           & 0.648 (0.071)            & 7.4            & 0.633 (0.078)            & 8.2            & 0.607 (0.079)            & 8.5            & 0.634 (0.078)          & 8.2           \\
ANCDE                             & 0.662 (0.083)          & 7.9           & 0.661 (0.083)            & \ul{7.1}      & 0.639 (0.080)            & 8.2            & 0.631 (0.073)            & 7.3            & 0.649 (0.080)          & 7.6           \\
EXIT                              & 0.595 (0.087)          & 10.1          & 0.580 (0.088)            & 10.5           & 0.578 (0.086)            & 10.1           & 0.564 (0.072)            & 10.4           & 0.579 (0.083)          & 10.3          \\
LEAP                              & 0.490 (0.062)          & 14.4          & 0.459 (0.070)            & 15.1           & 0.466 (0.074)            & 13.6           & 0.451 (0.074)            & 13.7           & 0.466 (0.070)          & 14.2          \\ \midrule
Neural Flow                       & 0.545 (0.059)          & 11.9          & 0.485 (0.067)            & 13.4           & 0.455 (0.058)            & 14.3           & 0.438 (0.054)            & 14.1           & 0.481 (0.059)          & 13.4          \\ \midrule
\textbf{Proposed method}          & \textbf{0.724 (0.090)} & \textbf{5.1}  & \textbf{0.720 (0.088)}   & \textbf{4.9}   & \textbf{0.691 (0.091)}   & \textbf{5.1}   & \textbf{0.697 (0.098)}   & \textbf{4.6}   & \textbf{0.708 (0.092)} & \textbf{4.9}  \\ \bottomrule
\end{tabular}
\caption{Average classification performance on 18 datasets under regular and three missing rates (Values in parentheses show the average of 18 standard deviations. The \textbf{top-ranking} and \ul{runner-up} methods are emphasized.) 
}\label{tab:result}
\end{table*}

\subsection{Robustness to Dataset Shift}
\paragraph{Descriptive Example. }
To provide comprehensive insights, we conducted a descriptive analysis of model performances using the `BasicMotions' dataset as a case study. We focused on five key methods: a standard RNN, GRU-ODE, ODE-LSTM, Neural CDE, Neural Flow, and our proposed method. Our investigation covered four different settings, including regular (0\% missingness), and three missing rates including 30\%, 50\%, and 70\%, respectively. 

\begin{figure}[ht]
    \centering\captionsetup{skip=5pt}
      \includegraphics[width=\linewidth]{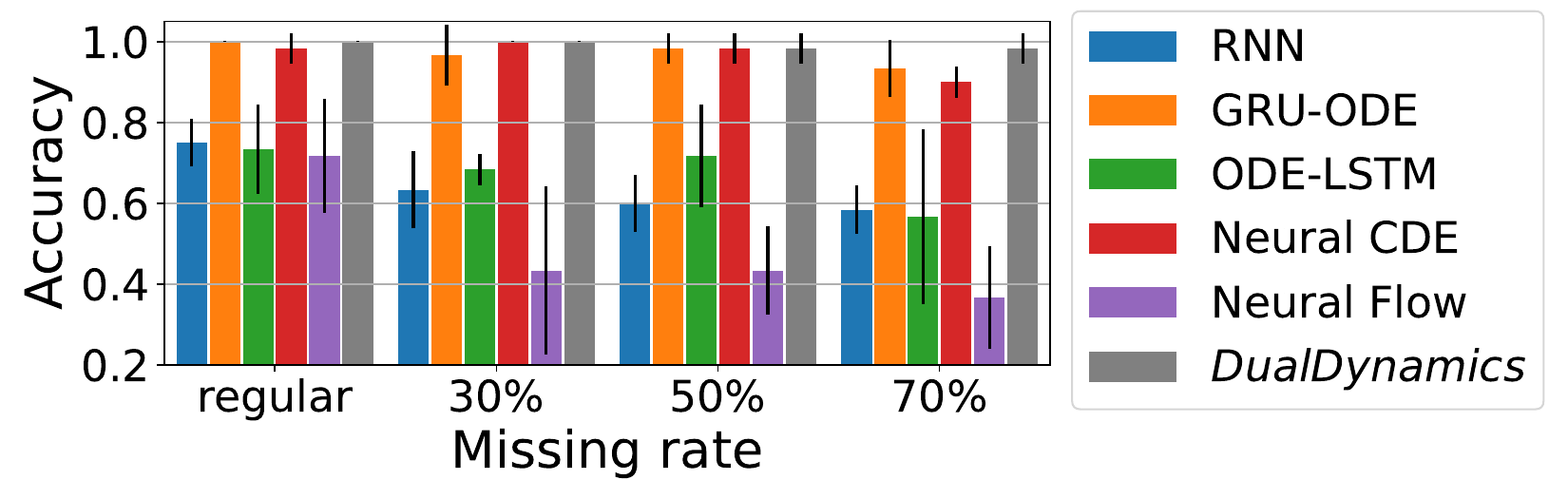}
    \caption{Classification performance comparison on the `BasicMotions' dataset with different scenarios}\label{fig:ex1}
\end{figure}
Figure~\ref{fig:ex1} illustrates the \textcolor{black}{classification performance} trend of each model concerning varying missing data rates. While the conventional RNN's performance decreases as the missing rate increases, GRU-ODE and ODE-LSTM outperform conventional RNNs but become less stability with increasing missing data. Neural CDE demonstrates superior performance, whereas Neural Flow shows performance worse than that of the conventional RNN.

\begin{figure}[ht]
    \centering\captionsetup{skip=5pt}
    \captionsetup[subfloat]{skip=5pt}
    \subfloat[Test loss without missingness (regular)]{
      \includegraphics[width=\linewidth]{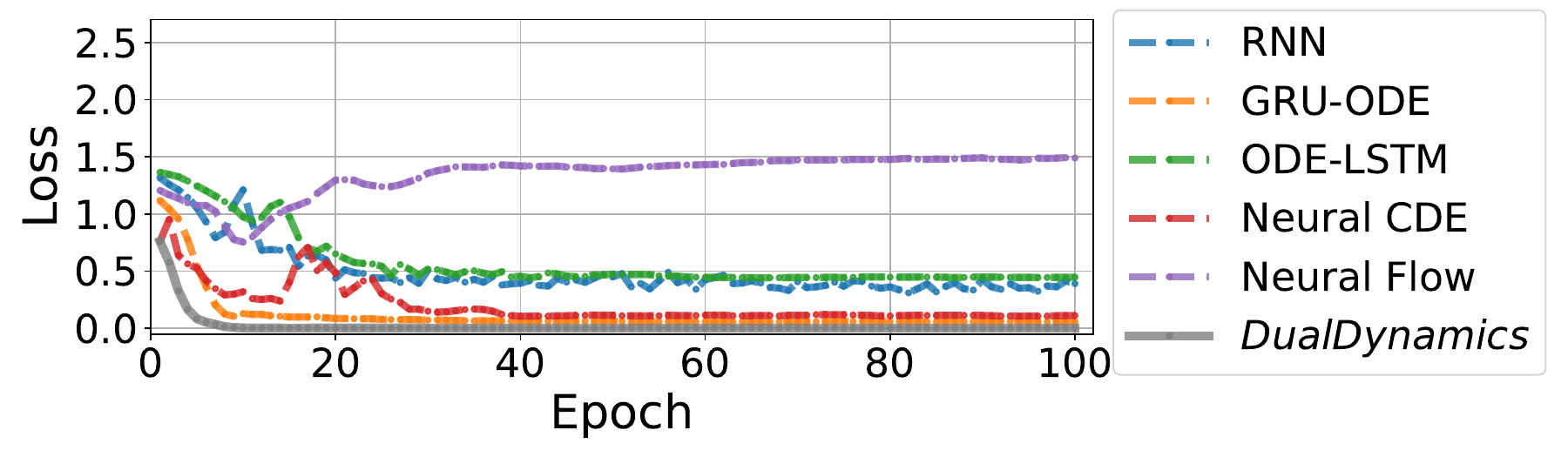}} \hfil
    \subfloat[Test loss with 50\% missing (irregular)]{
      \includegraphics[width=\linewidth]{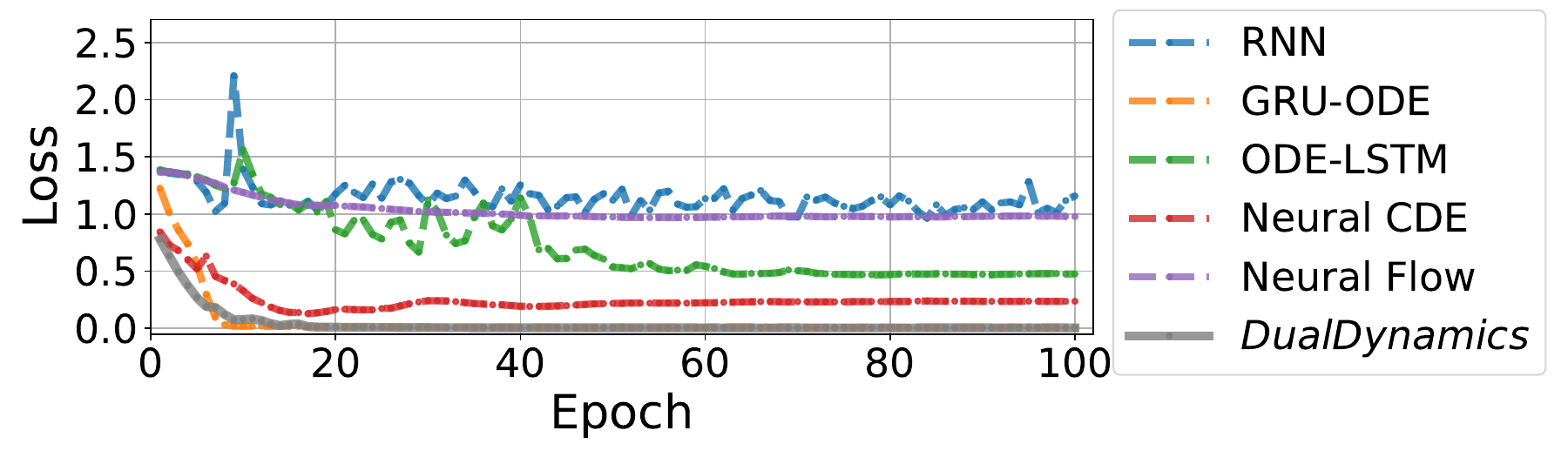}} 
    \caption{Stability comparison on the `BasicMotions` dataset, monitored over 100 epochs without early stopping}
    \label{fig:ex2}
\end{figure}
Our proposed method consistently demonstrates remarkable performance regardless of missing data rates. Further exploration is presented in Figure~\ref{fig:ex2}~(a) and (b), illustrating the test loss dynamics of the `BasicMotions' dataset, both with and without missingness. Despite its simplicity, Neural Flow shows limited convergence in both cases, whereas our proposed method achieves stability, outperforming both conventional approaches and Neural CDE. 
Furthermore, we included detailed results and computational time analysis in the supplementary material. 

\begin{table}[ht]
\scriptsize\centering\captionsetup{skip=5pt}
\begin{tabular}{@{}lcc@{}}
\toprule
\textbf{Configuration}                & \textbf{Accuracy} & \textbf{Loss} \\ \midrule
\textbf{Proposed method (Neural CDE)} & 0.708 (0.092)     & 0.686 (0.168) \\ \hdashline
-- Baseline                           & 0.667 (0.075)     & 0.765 (0.264) \\
-- Conventional MLP                   & 0.582 (0.087)     & 0.862 (0.148) \\
-- ResNet Flow                        & 0.651 (0.091)     & 0.777 (0.155) \\
-- GRU Flow                           & 0.660 (0.079)     & 0.768 (0.172) \\
-- Coupling Flow                      & 0.669 (0.097)     & 0.747 (0.186) \\ \midrule
\textbf{Proposed method (Neural ODE)} & 0.535 (0.066)     & 0.968 (0.072) \\ \hdashline
-- Baseline                           & 0.502 (0.069)     & 0.981 (0.078) \\
-- Conventional MLP                   & 0.504 (0.066)     & 0.980 (0.076) \\
-- ResNet Flow                        & 0.510 (0.060)     & 0.978 (0.077) \\
-- GRU Flow                           & 0.523 (0.071)     & 0.981 (0.080) \\
-- Coupling Flow                      & 0.523 (0.064)     & 0.970 (0.070) \\ \midrule
\textbf{Proposed method (Neural SDE)} & 0.538 (0.066)     & 0.968 (0.071) \\ \hdashline
-- Baseline                           & 0.509 (0.061)     & 0.978 (0.073) \\
-- Conventional MLP                   & 0.506 (0.069)     & 0.981 (0.077) \\
-- ResNet Flow                        & 0.520 (0.064)     & 0.969 (0.073) \\
-- GRU Flow                           & 0.521 (0.068)     & 0.981 (0.073) \\
-- Coupling Flow                      & 0.519 (0.068)     & 0.975 (0.079) \\ \bottomrule
\end{tabular}
\caption{Ablation study of the proposed method 
}\label{tab:ablation}
\end{table}

\begin{table*}[htbp]
\scriptsize\centering\captionsetup{skip=5pt}
\begin{tabular}{@{}lC{2.0cm}C{2.0cm}C{2.0cm}C{2.0cm}C{2.0cm}@{}}
\toprule
\multirow{2.5}{*}{\textbf{Methods}}             & \multicolumn{5}{c}{\textbf{Test MSE $\times 10^{-3}$}}                                                                                   \\ \cmidrule(l){2-6} 
                                              & \textbf{50\%}          & \textbf{60\%}          & \textbf{70\%}          & \textbf{80\%}          & \textbf{90\%}          \\ \midrule
RNN-VAE                                       & 13.418 ± 0.008         & 12.594 ± 0.004         & 11.887 ± 0.005         & 11.133 ± 0.007         & 11.470 ± 0.006         \\
L-ODE-RNN                                     & 8.132 ± 0.020          & 8.140 ± 0.018          & 8.171 ± 0.030          & 8.143 ± 0.025          & 8.402 ± 0.022          \\
L-ODE-ODE                                     & 6.721 ± 0.109          & 6.816 ± 0.045          & 6.798 ± 0.143          & 6.850 ± 0.066          & 7.142 ± 0.066          \\
mTAND-Full                                          & 4.139 ± 0.029          & 4.018 ± 0.048          & 4.157 ± 0.053          & 4.410 ± 0.149          & 4.798 ± 0.036          \\ \midrule
\textbf{Proposed method}                              & \textbf{3.631 ± 0.049} & \textbf{3.659 ± 0.028}          & \textbf{3.463 ± 0.032}    & \textbf{3.224 ± 0.044} & \textbf{3.114 ± 0.050} \\ \bottomrule
\end{tabular}
\caption{Performance of interpolation relative to observed percentage on the PhysioNet Mortality dataset 
}\label{tab:interpolation}
\end{table*}

\paragraph{Comprehensive Analysis. } Based on our initial findings, we extended our experiments to 18 public benchmark datasets from the the University of East Anglia (UEA) and the University of California Riverside (UCR) Time Series Classification Repository~\citep{bagnall2018uea,dau2019ucr}, encompassing Motion \& Human Activity Recognition (HAR), Electrocardiogram (ECG) \& Electroencephalogram (EEG), and Sensor domains. 
Following \citet{kidger2020neural} and \citet{oh2024stable}, we generated subsets with 30\%, 50\%, and 70\% missingness rates, creating four distinct scenarios. Data was split 70:15:15 for training, validation, and testing. We employed five repetition, evaluating mean performance and ranking across iterations. 

In the results presented in Table~\ref{tab:result}, our method distinctly stands out, demonstrating superiority when compared with other contemporary techniques. Considering the volatile nature of accuracy scores across diverse datasets, we prioritize rank statistics as a more consistent metric for comparative analysis. Across the four missing rate scenarios considered, our proposed method consistently achieves top-tier accuracy, confirming its robustness and effectiveness.

We conducted an in-depth comparative analysis of the classification efficacy and cross-entropy loss across various flow model designs, as detailed in Table~\ref{tab:ablation}. For implicit component, Neural CDE outperformed compared to Neural ODE and Neural SDE. The proposed method selected the flow configuration from ResNet flow, GRU flow, and Coupling flow during hyperparameter tuning. 
\textcolor{black}{For the comparison, we employed conventional MLP, which notably failed to meet specific flow criteria.}
Empirical evaluations indicate that the performance of conventional MLP is worse compared to the flow-based approach. This observation confirms the superiority of our proposed method.

\subsection{Interpolation of Missing Data}

The interpolation experiment utilized the 2012 PhysioNet Mortality dataset~\cite{silva2012predicting}, comprising 37 variables from ICU records over 48 hours post-admission. Following~\citet{rubanova2019latent}, timestamps were adjusted to the nearest minute, yielding up to 2880 intervals per series. We adhered to~\citet{shukla2021multi}'s methodology, varying observation frequency from 50\% to 90\% to predict remaining samples using 8000 instances.

In our interpolation experiments, we followed the experimental protocol advocated by \citet{shukla2021multi}. This strategy involved generating interpolated values based on a selected subset of data points to predict values for the unobserved time intervals. The interpolation spanned an observational range from 50\% to 90\% of the total data points. During the test phase, models were designed to interpolate values for the remaining intervals in the test set. The efficacy of the models was quantitatively assessed using Mean Squared Error (MSE), supported by five cross-validations.

Table~\ref{tab:interpolation} demonstrates that the interpolation performance of our proposed methods is superior compared to the benchmark models across all three flow configurations. The flow configuration achieved lower MSE errors compared to the non-flow neural network, a consistent and robust observation across various levels of observed data, ranging from 50\% to 90\%. Detailed experiment protocols and additional results are included in the supplementary material.

\subsection{Classification of Irregularly-Sampled Data}

We utilized the PhysioNet Sepsis dataset~\citep{reyna2019early}, comprising 40,335 patient instances with 34 temporal variables. Following~\citet{kidger2020neural}, we addressed data irregularity using two classification approaches: with and without observation intensity (OI). Observation intensity (OI) indicates patient condition severity, augmenting each time series point with an intensity index when utilized. Model performance was evaluated using Area Under the Receiver Operating Characteristic Curve (AUROC), with benchmark comparisons from~\citet{jhin2022exit}.

\begin{table}[htbp]
\scriptsize\centering\captionsetup{skip=5pt}
\begin{tabular}{@{}lcccc@{}} 
\toprule
\multirow{2.5}{*}{\textbf{Methods}} & \multicolumn{2}{c}{\textbf{Test AUROC}} & \multicolumn{2}{c}{\textbf{Memory (MB)}} \\ \cmidrule(lr){2-3}\cmidrule(lr){4-5} 
\multicolumn{1}{c}{}  & \multirow{1}{*}{\textbf{OI}}  & \multirow{1}{*}{\textbf{No OI}}  & \multirow{1}{*}{\textbf{OI}} & \multirow{1}{*}{\textbf{No OI}} \\ \midrule
GRU-$\Delta t$ & 0.878 ± 0.006 & 0.840 ± 0.007 & 837 & 826  \\
GRU-D & 0.871 ± 0.022 & 0.850 ± 0.013 & 889 & 878  \\
GRU-ODE & 0.852 ± 0.010 & 0.771 ± 0.024 & 454 & 273  \\
ODE-RNN & 0.874 ± 0.016 & 0.833 ± 0.020 & 696 & 686  \\
Latent-ODE & 0.787 ± 0.011 & 0.495 ± 0.002 & 133 & 126  \\
ACE-NODE & 0.804 ± 0.010 & 0.514 ± 0.003 & 194 & 218  \\
Neural CDE  & 0.880 ± 0.006 & 0.776 ± 0.009 & 244 & 122 \\
ANCDE & 0.900 ± 0.002 & 0.823 ± 0.003 & 285 & 129 \\ 
EXIT & 0.913 ± 0.002 & 0.836 ± 0.003 & 257 & 127 \\ \midrule
\textbf{Proposed method} & \textbf{0.918 ± 0.003} & \textbf{0.873 ± 0.004} & 453 & 233 \\ 
\bottomrule
\end{tabular}
\caption{AUROC on PhysioNet Sepsis dataset}\label{tab:sepsis}
\end{table}

In Table~\ref{tab:sepsis}, our method demonstrates superior AUROC scores in both scenarios. Consistent with prior experimental findings, models based on flow exhibit enhanced efficacy relative to their non-flow neural network counterparts. These results confirm the effectiveness of our proposed methodologies in addressing challenges associated with time series irregularity and missingness.

\subsection{Forecasting with the Partial Observations}
We utilized two datasets, MuJoCo and Google, with different levels of observation for the forecasting task. Detailed explanations of each dataset, implementation details, and results are provided in the supplementary material.

\paragraph{MuJoCo.} 
We utilized the MuJoCo dataset~\citep{todorov2012mujoco}, based on the Hopper configuration from the DeepMind control suite~\citep{tassa2018deepmind}. Following~\citet{jhin2023learnable,jhin2024attentive}, we used 50 initial points to predict the next 10, introducing 30\%, 50\%, and 70\% missing data scenarios.
Table~\ref{tab:mujoco} illustrates the forecasting performance across diverse conditions, with our method consistently demonstrating minimal MSE in all settings. 
\begin{table}[ht]
\scriptsize\centering\captionsetup{skip=5pt}
\begin{tabular}{@{}lC{1.0cm}C{1.0cm}C{1.0cm}C{1.0cm}@{}}
\toprule
\multirow{3.5}{*}{\textbf{Methods}} & \multicolumn{4}{c}{\textbf{Test MSE}} \\ \cmidrule(lr){2-5}
                                  & \textbf{Regular}       & \textbf{30\% Dropped}  & \textbf{50\% Dropped}  & \textbf{70\% Dropped}  \\ \midrule
GRU-$\Delta t$                    & \makecell{0.223 \\ ± 0.020} & \makecell{0.198 \\ ± 0.036} & \makecell{0.193 \\ ± 0.015} & \makecell{0.196 \\ ± 0.028} \\
GRU-D                             & \makecell{0.578 \\ ± 0.042} & \makecell{0.608 \\ ± 0.032} & \makecell{0.587 \\ ± 0.039} & \makecell{0.579 \\ ± 0.052} \\
GRU-ODE                           & \makecell{0.856 \\ ± 0.016} & \makecell{0.857 \\ ± 0.015} & \makecell{0.852 \\ ± 0.015} & \makecell{0.861 \\ ± 0.015} \\
ODE-RNN                           & \makecell{0.328 \\ ± 0.225} & \makecell{0.274 \\ ± 0.213} & \makecell{0.237 \\ ± 0.110} & \makecell{0.267 \\ ± 0.217} \\
Latent-ODE                        & \makecell{0.029 \\ ± 0.011} & \makecell{0.056 \\ ± 0.001} & \makecell{0.055 \\ ± 0.004} & \makecell{0.058 \\ ± 0.003} \\
Augmented-ODE                     & \makecell{0.055 \\ ± 0.004} & \makecell{0.056 \\ ± 0.004} & \makecell{0.057 \\ ± 0.005} & \makecell{0.057 \\ ± 0.005} \\
ACE-NODE                          & \makecell{0.039 \\ ± 0.003} & \makecell{0.053 \\ ± 0.007} & \makecell{0.053 \\ ± 0.005} & \makecell{0.052 \\ ± 0.006} \\
Neural CDE                        & \makecell{0.028 \\ ± 0.002} & \makecell{0.027 \\ ± 0.000} & \makecell{0.027 \\ ± 0.001} & \makecell{0.026 \\ ± 0.001} \\
ANCDE                             & \makecell{0.026 \\ ± 0.001} & \makecell{0.025 \\ ± 0.001} & \makecell{0.025 \\ ± 0.001} & \makecell{0.024 \\ ± 0.001} \\
EXIT                              & \makecell{0.026 \\ ± 0.000} & \makecell{0.025 \\ ± 0.004} & \makecell{0.026 \\ ± 0.000} & \makecell{0.026 \\ ± 0.001} \\
LEAP                              & \makecell{0.022 \\ ± 0.002} & \makecell{0.022 \\ ± 0.001} & \makecell{0.022 \\ ± 0.002} & \makecell{0.022 \\ ± 0.001} \\ \midrule
\textbf{Proposed method}             & \makecell{\textbf{0.006} \\ \textbf{± 0.001}} & \makecell{\textbf{0.008} \\ \textbf{± 0.001}} & \makecell{\textbf{0.008} \\ \textbf{± 0.001}} & \makecell{\textbf{0.008} \\ \textbf{± 0.001}} \\ \bottomrule
\end{tabular}
\caption{Forecasting performance comparison on the MuJoCo dataset with varying percentages of data dropped
}\label{tab:mujoco}
\bigskip
    \centering\captionsetup{skip=5pt}
    \captionsetup[subfloat]{skip=5pt}
          \includegraphics[width=\linewidth]{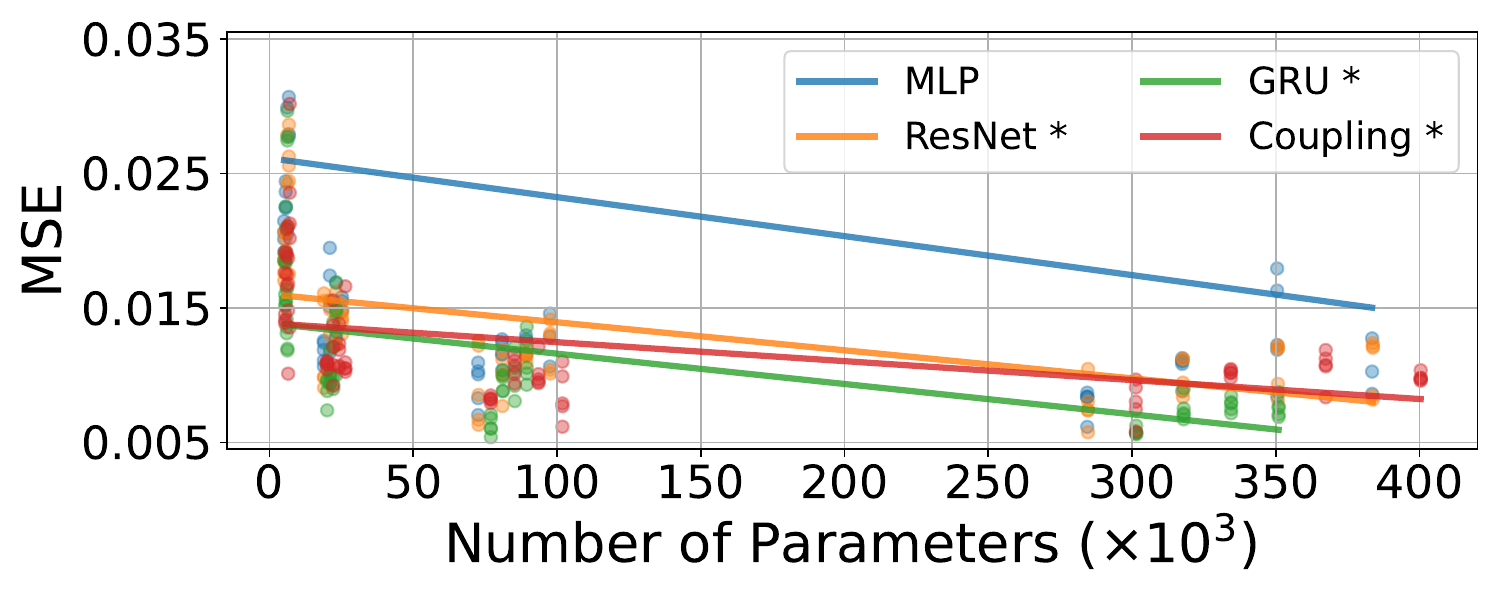}
    \captionof{figure}{Forecasting performance versus the number of parameters on MuJoCo dataset with regular scenario (different flow model (with remarked $*$) versus conventional MLP)}
    \label{fig:memory_mujoco}
\end{table}

\paragraph{Google.} 
We used the Google stock data (2011-2021)\citep{jhin2022exit}, forecasting 5 price metrics for the next 10 days based on 50 days of history. Following \citet{jhin2022exit}'s protocol, we introduced 30\%, 50\%, and 70\% missing data scenarios.
Table~\ref{tab:google} presents the outcomes of forecasting stock data with partial observations. When comparing with benchmark methods, our method consistently presents lower MSE in the four different scenarios. 
\begin{table}[htbp]
\scriptsize\centering\captionsetup{skip=5pt}
\begin{tabular}{@{}lC{1.0cm}C{1.0cm}C{1.0cm}C{1.0cm}@{}}
\toprule
\multirow{3.5}{*}{\textbf{Methods}} & \multicolumn{4}{c}{\textbf{Test MSE}} \\ \cmidrule(lr){2-5}
                                  & \textbf{Regular}         & \textbf{30\% Dropped}    & \textbf{50\% Dropped}    & \textbf{70\% Dropped}    \\ \midrule
GRU-$\Delta t$                    & \makecell{0.0406 \\ ± 0.002} & \makecell{0.0043 \\ ± 0.003} & \makecell{0.0041 \\ ± 0.002} & \makecell{0.0041 \\ ± 0.001} \\
GRU-D                             & \makecell{0.0040 \\ ± 0.004} & \makecell{0.0058 \\ ± 0.018} & \makecell{0.0039 \\ ± 0.001} & \makecell{0.0040 \\ ± 0.001} \\
GRU-ODE                           & \makecell{0.0028 \\ ± 0.016} & \makecell{0.0029 \\ ± 0.001} & \makecell{0.0031 \\ ± 0.004} & \makecell{0.0030 \\ ± 0.002} \\
ODE-RNN                           & \makecell{0.0311 \\ ± 0.044} & \makecell{0.0318 \\ ± 0.066} & \makecell{0.0322 \\ ± 0.056} & \makecell{0.0324 \\ ± 0.053} \\
Latent-ODE                        & \makecell{0.0030 \\ ± 0.011} & \makecell{0.0032 \\ ± 0.001} & \makecell{0.0033 \\ ± 0.002} & \makecell{0.0032 \\ ± 0.003} \\
Augmented-ODE                     & \makecell{0.0023 \\ ± 0.002} & \makecell{0.0025 \\ ± 0.002} & \makecell{0.0029 \\ ± 0.006} & \makecell{0.0023 \\ ± 0.005} \\
ACE-NODE                          & \makecell{0.0022 \\ ± 0.003} & \makecell{0.0024 \\ ± 0.001} & \makecell{0.0022 \\ ± 0.002} & \makecell{0.0025 \\ ± 0.002} \\
Neural CDE                        & \makecell{0.0037 \\ ± 0.062} & \makecell{0.0038 \\ ± 0.073} & \makecell{0.0032 \\ ± 0.035} & \makecell{0.0039 \\ ± 0.038} \\
ANCDE                             & \makecell{0.0020 \\ ± 0.002} & \makecell{0.0022 \\ ± 0.001} & \makecell{0.0021 \\ ± 0.002} & \makecell{0.0021 \\ ± 0.001} \\
EXIT                              & \makecell{0.0016 \\ ± 0.002} & \makecell{0.0016 \\ ± 0.001} & \makecell{0.0017 \\ ± 0.001} & \makecell{0.0019 \\ ± 0.003} \\ \midrule
\textbf{Proposed method}                 & \makecell{\textbf{0.0010} \\ \textbf{± 0.0001}} & \makecell{\textbf{0.0010} \\ \textbf{± 0.0001}} & \makecell{\textbf{0.0009} \\ \textbf{± 0.0001}} & \makecell{\textbf{0.0013} \\ \textbf{± 0.0002}} \\ \bottomrule
\end{tabular}
\caption{Forecasting performance comparison on the Google dataset with varying percentages of data dropped
}\label{tab:google}
\bigskip
    \centering\captionsetup{skip=5pt}
    \captionsetup[subfloat]{skip=5pt}
          \includegraphics[width=\linewidth]{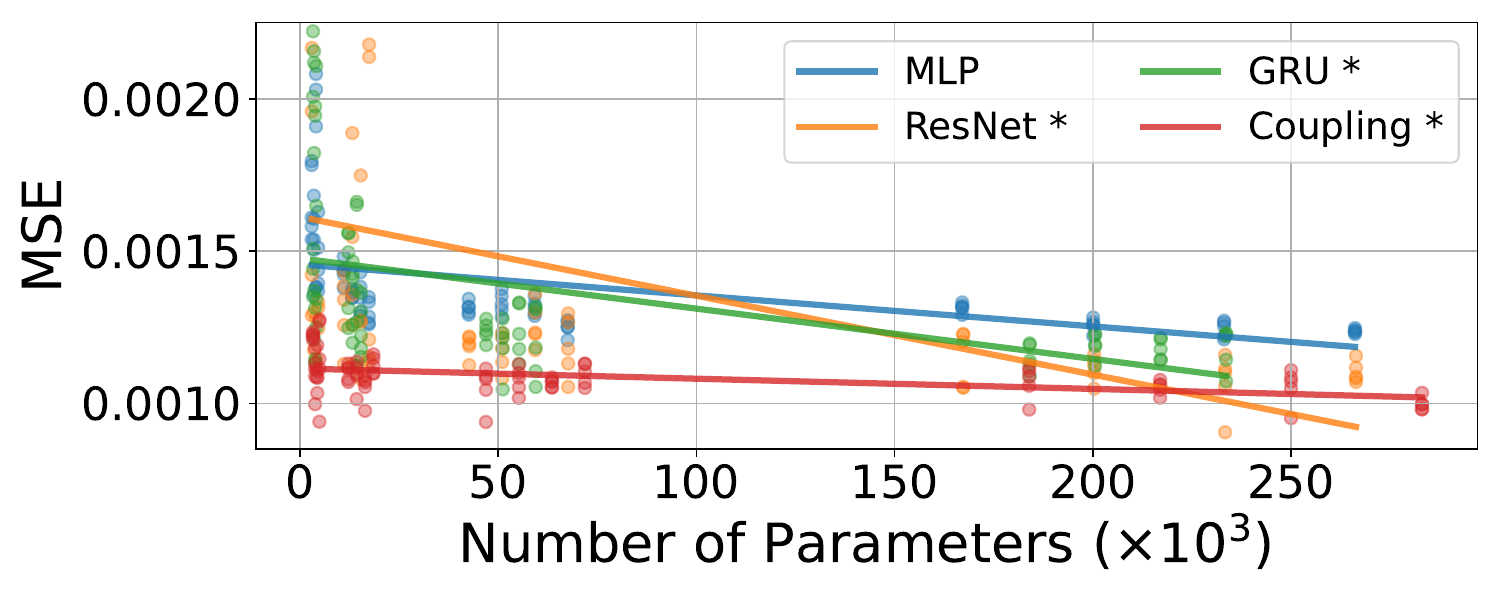}
    \captionof{figure}{Forecasting performance versus the number of parameters on Google dataset with regular scenario (different flow model (with remarked $*$) versus conventional MLP)}
    \label{fig:memory_google}
\end{table}

\section{Discussion}
\paragraph{Performance versus Network Capacity. }
Our analysis reveals that the performance improvements of \texttt{DualDynamics} are not merely a result of increased network capacity, but rather stem from the synergistic integration of implicit and explicit methods in our framework. 

To validate \texttt{DualDynamics}' effectiveness, we conducted controlled experiments comparing it against models with similar parameter counts. Figures~\ref{fig:memory_mujoco} and \ref{fig:memory_google} illustrate the relationship between parameter count and MSE for our flow architecture versus conventional MLPs. 
Figure~\ref{fig:memory_mujoco} shows that increasing MLP capacity yields limited improvement on the MuJoCo dataset, while \texttt{DualDynamics} maintains superior performance. Figure~\ref{fig:memory_google} demonstrates that with optimal flow configuration, \texttt{DualDynamics} consistently outperforms MLPs on the Google dataset across all model sizes.

This performance advantage persists even when adjusting baseline model complexity to match \texttt{DualDynamics}, indicating that our framework's superiority stems from its architectural design rather than network capacity. These results highlight \texttt{DualDynamics}' efficiency in capturing complex temporal dynamics in irregular time series.

\begin{figure}[htbp]
    \centering\captionsetup{skip=5pt}
    \captionsetup[subfloat]{skip=5pt}
    \subfloat[Physionet Mortality]{
      \includegraphics[width=\linewidth]{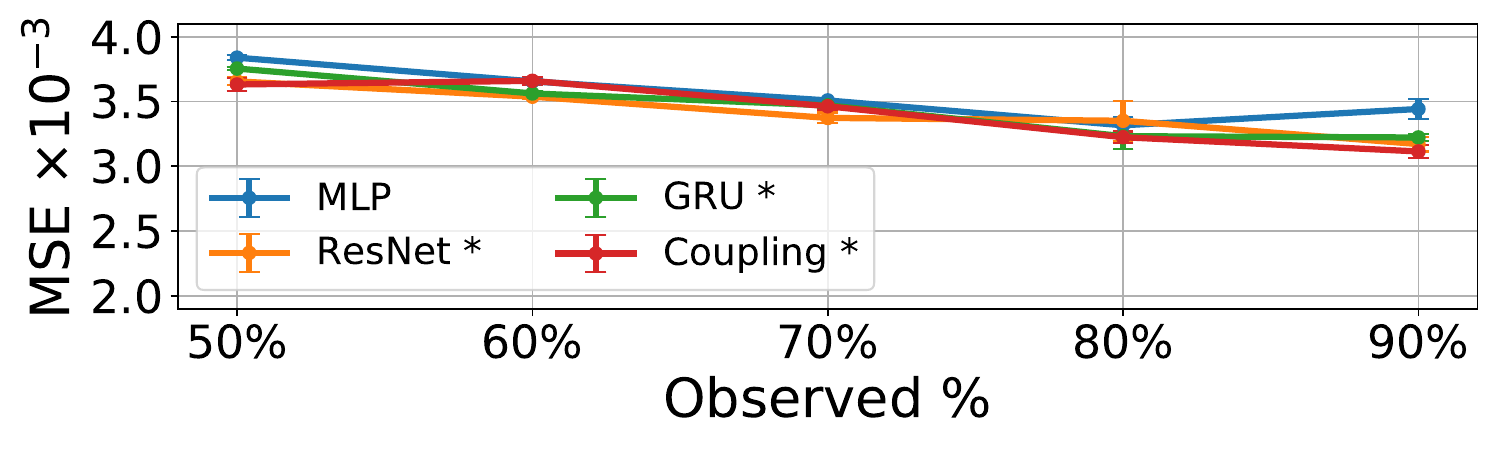}} \\ 
    \subfloat[Physionet Sepsis]{
      \includegraphics[width=\linewidth]{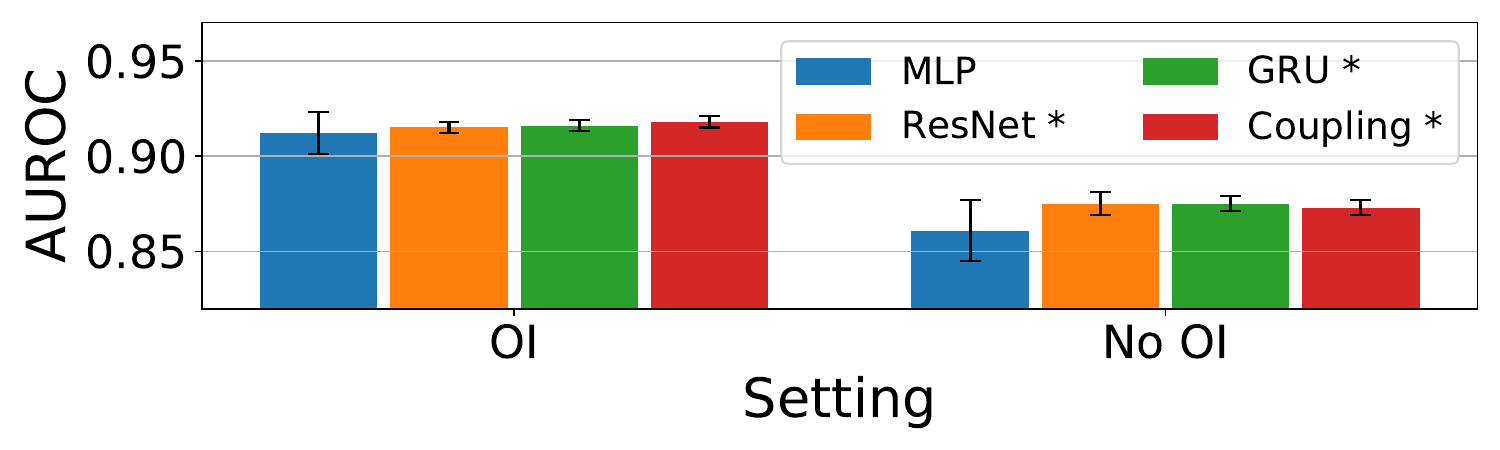}} \\
    \subfloat[MuJoCo]{
      \includegraphics[width=\linewidth]{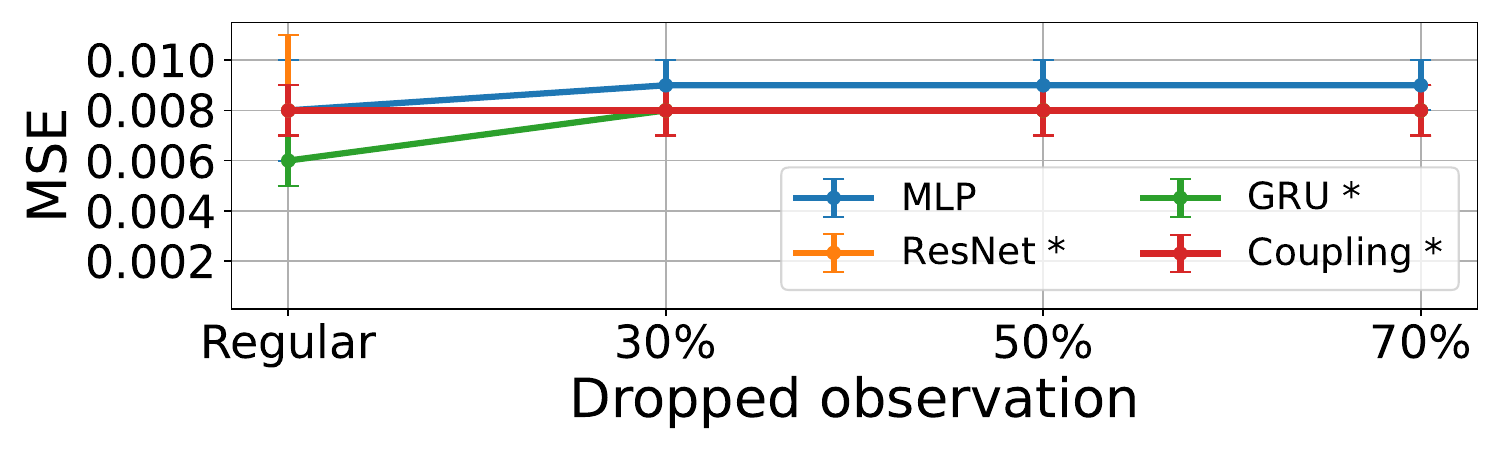}} \\
    \subfloat[Google]{
      \includegraphics[width=\linewidth]{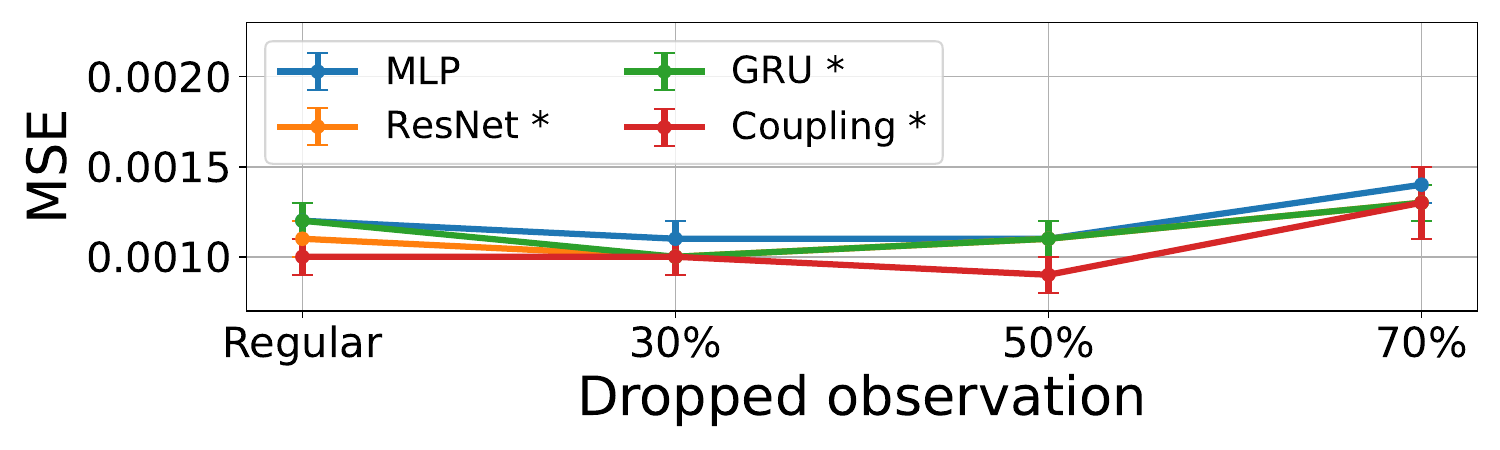}}
    \caption{Ablation study of different flow model (with remarked $*$) versus conventional MLP}
    \label{fig:summary}
\end{figure}

\paragraph{Ablation Study of Flow Model Configurations. }
Figure~\ref{fig:summary} presents a comprehensive overview of our ablation study with four datasets, illustrating the performance of various flow configurations across different tasks, as detailed in the preceding sections. The results demonstrate a clear advantage of flow-based architectures over the conventional MLP approach, refer to the detailed results in the supplementary material. 
These results suggest that incorporating implicit and explicit could lead to significant improvements in model performance and capabilities.

\paragraph{Variations in Our Framework. }
Our framework's architecture allows for several potential variations, each with distinct trade-offs. 
One possible modification is to utilize only $\vz(t)$ instead of $\hat{\vz}(t)$ after training the unified framework. However, this approach significantly limits the model's expressiveness, as it fails to leverage the unique properties of the flow model. 
Another consideration is the replacement of the flow model with more sophisticated architectures, such as feedforward networks~\citep{hornik1989multilayer}, gated recurrent units~\citep{chung2014empirical}, temporal convolutional networks~\citep{lea2017temporal}, or transformer-based attention mechanisms~\citep{vaswani2017attention}. 
For further details and quantitative comparisons, we refer the reader to the supplementary material.

\section{Conclusion}\label{Conclusion}

We introduce \texttt{DualDynamics}, an innovative methodology that integrates implicit and explicit methods for modeling irregularly-sampled time series data. This sophisticated paradigm excels in capturing intricate temporal characteristics inherent in continuous-time processes, offering an enhanced framework for representation learning in time series analysis. By synergistically combining these approaches, our method strikes a balance between expressive power and computational efficiency, addressing key challenges in temporal modeling for the irregular time series.

Our approach consistently surpasses existing benchmarks in classification, interpolation, and forecasting tasks, significantly elevating the precision and depth of understanding in temporal dynamics. 
These outstanding results highlight \texttt{DualDynamics}' potential for diverse real-world applications. Its adaptability to various temporal modeling challenges positions it as a significant advancement in the irregular time series analysis in practices.

%
\section*{Ethics Statement}
We commit to conducting our research with integrity, ensuring ethical practices and the responsible use of technology, fully aligning our efforts with established academic and scientific standards to promote trust and accountability.
%
%
\section*{Acknowledgments}
We thank the teams and individuals for their efforts in the real-world dataset preparation and curation for our research, especially the UEA \& UCR repository for the numerous datasets that we extensively analyzed.

This research was supported by Basic Science Research Program through the National Research Foundation of Korea (NRF) funded by the Ministry of Education (RS-2024-00407852); Korea Health Technology R\&D Project through the Korea Health Industry Development Institute (KHIDI), funded by the Ministry of Health and Welfare, Republic of Korea (HI19C1095); the Institute of Information \& communications Technology Planning \& Evaluation (IITP) grant funded by the Korea government (MSIT) (No.2020-0-01336, Artificial Intelligence Graduate School Program (UNIST)); and the Institute of Information \& communications Technology Planning \& Evaluation (IITP) grant funded by the Korea government (MSIT) (No.RS-2021-II212068, Artificial Intelligence Innovation Hub).

{\small
\bibliography{references}
}

\newpage
\onecolumn
\appendix

\section{Experimental Details}~\label{appendix:experiment}

We consider following experiments: (1) Robustness to Dataset Shift, (2) Classification of irregularly-sampled data, (3) Interpolation of Missing Data, and (4) Forecasting with the partial observations. 
All experiments were performed using a server on Ubuntu 22.04 LTS, equipped with an Intel(R) Xeon(R) Gold 6242 CPU and multiple NVIDIA A100 40GB GPUs. The source code for our experiments and datasets can be accessed at \url{https://github.com/yongkyung-oh/DualDynamics}.

\begin{figure*}[htbp]
    \centering\captionsetup{skip=5pt}
    \includegraphics[scale=0.40]{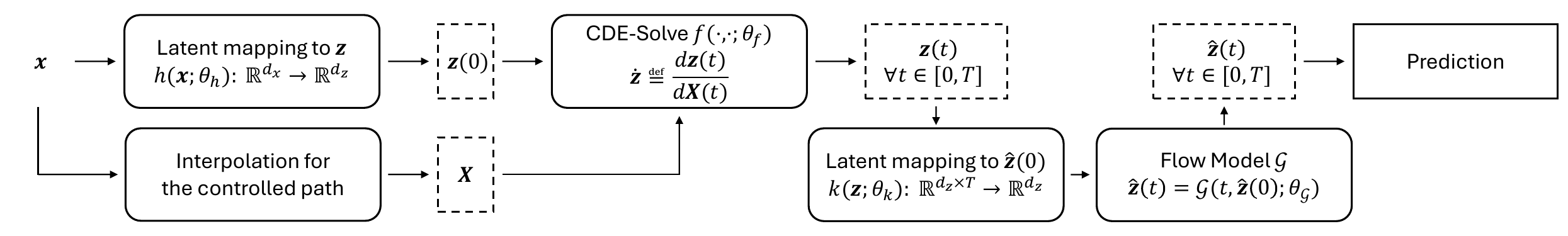}
    \caption{The detailed architecture of the proposed \texttt{DualDynamics} with Neural CDE 
    }\label{fig:detail}
\end{figure*}

Figure~\ref{fig:detail} illustrates the architecture of our proposed \texttt{DualDynamics}. This framework integrates implicit and explicit learning components for enhanced time series modeling. The implicit component can be implemented using Neural ODEs, Neural CDEs, or Neural SDEs, each adept at capturing complex temporal dependencies. The explicit component, designed to refine these representations, utilizes advanced neural flow models: ResNet Flow, GRU Flow, or Coupling Flow. This combination of implicit and explicit methods in \texttt{DualDynamics} aims to balance flexibility in modeling irregular time series with computational efficiency, potentially improving performance. 

\subsection{Implicit Component: NDE-Based Approaches}
\subsubsection{Neural ODEs}
Let $\vx = (x_0, x_1, \ldots, x_n)$ be a vector of original (possibly, irregularly-sampled) observations. For a given time $t$, let us denote a latent state as $\vz(t)$. Neural ODEs can be formally expressed as:
\begin{align}\label{eq:neural_ode_}
\vz(t) & = \vz(0) + \int_0^t f(s,\vz(s);\theta_f) \rd s, 
\end{align}
with the initial condition $\vz(0) = h(\vx;\theta_{h})$ where $h: \mathbb{R}^{d_x} \rightarrow \mathbb{R}^{d_z}$ is an affine function with parameter $\theta_{h}$. In \Eqref{eq:neural_ode_}, the function $h(\cdot;\theta_{h})$ can capture the intrinsic features from the input to the latent state.
$f(s,\vz(s);\theta_f)$, which approximates the rate of change, $\dot{\vz} \equiv \frac{\rd \vz(t)}{\rd t}$, is modelled using a neural network parameterized by $\theta_f$. However, the performance of Neural ODEs often diminishes due to their dependence on initial conditions~\citep{kidger2022neural}.
\subsubsection{Neural CDEs}
Neural CDEs provide a framework for constructing a continuous-time representation by formulating a control path $X(t)$, generated through natural cubic spline interpolation of observed data~\citep{kidger2020neural}. The Neural CDEs is expressed via the Riemann--Stieltjes integral:
\begin{align}\label{eq:neural_cde_}
\vz(t) & = \vz(0) + \int_0^t f(s,\vz(s); \theta_f) \, \rd X(s),  
\end{align}
with the initial condition $\vz(0) = h(x_0;\theta_{h})$ where $f(s, \vz(s); \theta_f)$ provides an approximation to $\frac{\rd \vz(t)}{\rd X(t)}$, differentiating Neural CDEs from Neural ODEs in their approximation approach. 
To evaluate the integral in \Eqref{eq:neural_cde_}, one may utilize the conventional ODE solvers by recognizing that $\dot{\vz}(t) \equiv \frac{\rd \vz(t)}{\rd X(t)}$.

\subsubsection{Neural SDEs}
Neural SDEs extend the concept of Neural ODEs by incorporating a stochastic term to model uncertainty and noise in the dynamics~\cite{oh2024stable}. For a given time $t$ and latent state $\vz(t)$, Neural SDEs can be formally expressed as:
\begin{align}\label{eq:neural_sde_}
\vz(t) & = \vz(0) + \int_0^t f(s,\vz(s);\theta_f) \rd s + \int_0^t g(s,\vz(s);\theta_g) \rd W(s),
\end{align}
with the initial condition $\vz(0) = h(\vx;\theta_{h})$ where $h: \mathbb{R}^{d_x} \rightarrow \mathbb{R}^{d_z}$ is an affine function with parameter $\theta_{h}$, similar to Neural ODEs.
Here, $f(t,\vz(t);\theta_f)$ represents the drift term, which models the deterministic part of the dynamics, while $g(t,\vz(t);\theta_g)$ is the diffusion term that captures the stochastic component. $W(t)$ denotes a standard Wiener process (or Brownian motion). In \Eqref{eq:neural_sde_}, both $f$ and $g$ are modeled using neural networks parameterized by $\theta_f$ and $\theta_g$, respectively.
Neural SDEs provide a framework for modeling complex, stochastic temporal dynamics, allowing for the representation of uncertainty in the latent state evolution as continues-time perspective.

\subsection{Explicit Component: Flow Model Configurations}\label{appendix:flow}
We integrate three distinct flow architectures proposed in \citet{bilovs2021neural} within the proposed framework.

\paragraph{ResNet Flow. } Originating from the concept of Residual Networks (ResNets) \cite{he2016deep}, the ResNet flow model can be conceptualized as a continuous-time analogue of ResNets. The core dynamics of $\vz(t)$ in ResNet flow are described by the equation:
\begin{align}
    \gG(t,\vz) = \vz + \varphi(t)g(t,\vz),
\end{align}
where $\varphi$ represents a temporal embedding function and $g$ denotes a continuous, nonlinear transformation parameterized by $\theta_g$. To address the inherent non-invertibility in standard ResNets, spectral normalization techniques \cite{gouk2021regularisation} are employed, ensuring a bounded Lipschitz constant and thus, invertibility \cite{behrmann2019invertible}.

\paragraph{GRU Flow. } The GRU flow model extends the Gated Recurrent Unit (GRU) architecture \cite{chung2014empirical} to a continuous-time setting. In this model, the hidden state $\vz$ evolves continuously according to an ordinary differential equation, akin to the updates in traditional discrete GRUs. \citet{de2019gru} introduced GRU-ODE, and \citet{bilovs2021neural} further refined it to an invertible form within the Neural Flow context, as expressed by:
\begin{align}
    \gG(t, \vz) = \vz + \varphi(t) (1 - g_1(t, \vz)) \odot (g_2(t, \vz) - \vz),
\end{align}
where $\varphi(t)$ is a time-dependent embedding function, and $g_1$, $g_2$ are adapted from the GRU, where 
$g_1(t,\vz)=\alpha \cdot \texttt{sigmoid}(f_1(t,\vz))$, and $g_2(t,\vz) = \texttt{tanh}(f_2(t,g_3(t, \vz) \odot \vz)$, and $g_3(t,\vz)=\beta \cdot \texttt{sigmoid}(f_3(t,\vz)$. $f_1$, $f_2$, $f_3$ are any arbitrary neural networks.

\paragraph{Coupling Flow. } The concept of Coupling flow, initially introduced by \citet{dinh2014nice} and \citet{dinh2016density}, involves a bijective transformation characterized by its computational efficiency and flexibility. Input dimensions are partitioned into two disjoint subsets $d_1$ and $d_2$, leading to the transformation:
\begin{align}
    \gG(t)_{d_1} = \vz_{d_1} \exp(u(t, \vz_{d_2}) \varphi_u(t)) + v(t, \vz_{d_2}) \varphi_v(t),
\end{align}
with $u$ and $v$ as neural networks, and $\varphi_u$, $\varphi_v$ as time-dependent embedding functions. Two disjoint set $d_1$ and $d_2$ satisfies $d_1 \bigcup d_2 =\left\{1,2,\ldots,d_z\right\}$. 
This transformation is inherently invertible, maintaining the topological properties of the state space.

Each flow model exhibits its own strengths and limitations. The choice of a flow model should be contingent upon the dataset's nature, the problem's specific requirements, and computational constraints. Therefore, the selection of a flow model is empirically refined during hyperparameter tuning.


\subsection{Training Procedure}\label{appendix_algorithm}

We present the algorithm for the proposed \texttt{DualDynamics} framework. The presented algorithm \ref{alg:dualdynamics} concisely captures the training procedure of the \texttt{DualDynamics} framework. Here we showcase the example of proposed method using Neural CDE for classification task with cross-entropy loss.

\begin{algorithm}[H]
\small
\caption{Training Procedure for \texttt{DualDynamics} (using Neural CDE for classification task)}\label{alg:dualdynamics}
\begin{algorithmic}[1]
\State Initialized model parameters $\Theta = \{\theta_h, \theta_f, \theta_k, \theta_\gG, \theta_{MLP}\}$.

\State \textbf{Forward Pass:}
\State $\quad$ Compute initial latent state $\vz(0) = h(\vx_0; \theta_h)$ and solve Neural CDE to obtain $\vz(t)$:
\[
\vz(t) = \vz(0) + \int_0^t f(s, \vz(s); \theta_f) \, \rd X(s) 
\] 
\State $\quad$ Aggregate $\vz(t)$ to compute $\hat{\vz}(0) = k(\vz; \theta_k)$ and compute $\hat{\vz}(t)$ using flow model $\gG$:
\[
\hat{\vz}(t) = \gG(t, \hat{\vz}(0); \theta_\gG) 
\] 

\State \textbf{Loss Computation:}
\State $\quad$ Compute cross-entropy loss $\mathcal{L}$ using $\hat{\bm{z}}(T)$ and target $y$.
\[
\mathcal{L} = -\sum_{i=1}^C y_i \log(\text{softmax}(MLP(\hat{\bm{z}}(T); \theta_{MLP}))_i)
\]

\State \textbf{Backward Pass:}
\State $\quad$ Compute gradients $\nabla_{\theta} \mathcal{L}$ using adjoint-based backpropagation: 
\[
\frac{\partial \mathcal{L}}{\partial \theta_f} = \int_0^T \bm{\lambda}_{\vz}(t)^\top \frac{\partial f(t, \vz(t); \theta_f)}{\partial \theta_f} \, \frac{\rd X(t)}{\rd t} \rd t, \qquad
\frac{\partial \mathcal{L}}{\partial \theta_\gG} = \bm{\lambda}_{\hat{\vz}}(T)^\top \frac{\partial \gG(T, \hat{\vz}(0); \theta_{\gG})}{\partial \theta_\gG}.
\]

\State $\quad$ Update all parameters.
$\Theta \gets \Theta - \eta \nabla_{\Theta} \mathcal{L}$
\end{algorithmic}
\end{algorithm}


\section{Detailed Results for ``Robustness to Dataset Shift''}\label{appendix:robust}

\subsection{Data Preparation}
We conducted classification experiments across 18 diverse datasets, in three distinct domains, from the University of East Anglia (UEA) and the University of California Riverside (UCR) Time Series Classification Repository \footnote{\url{http://www.timeseriesclassification.com/}}~\citep{bagnall2018uea,dau2019ucr} using the python library \texttt{sktime}~\citep{loning2019sktime}, as outlined in Table~\ref{tab:data}. 
Addressing the challenge posed by varying lengths of time series, the strategy of uniform scaling was employed to align all series to the dimension of the longest series~\citep{keogh2003efficiently,yankov2007detecting,gao2018efficient,tan2019time}. 
The partitioning of the dataset into training, validation, and testing subsets was executed in proportions of 0.70, 0.15, and 0.15 respectively. 
Subsequent to this partitioning, a random assortment of missing observations was introduced for each variable. Lastly, these modified variables are integrated into combined dataset. 

\begin{table*}[htbp]
\scriptsize\centering\captionsetup{skip=5pt}
\begin{tabular}{@{}clrrrr@{}}
\toprule
\textbf{Domain}                         & \multicolumn{1}{c}{\textbf{Dataset}} & \multicolumn{1}{c}{\textbf{\begin{tabular}[c]{@{}c@{}}Total number of \\ samples\end{tabular}}} & \multicolumn{1}{c}{\textbf{\begin{tabular}[c]{@{}c@{}}Number of \\ classes\end{tabular}}} & \multicolumn{1}{c}{\textbf{\begin{tabular}[c]{@{}c@{}}Dimension of \\ time series\end{tabular}}} & \multicolumn{1}{c}{\textbf{\begin{tabular}[c]{@{}c@{}}Length of \\ time series\end{tabular}}} \\ \midrule
\multirow{6}{*}{\textbf{Motion \& HAR}} & \textbf{BasicMotions}                & 80                                                   & 4                                              & 6                                                     & 100                                                \\
                                        & \textbf{Epilepsy}                    & 275                                                  & 4                                              & 3                                                     & 206                                                \\
                                        & \textbf{PickupGestureWiimoteZ}       & 100                                                  & 10                                             & 1                                                     & 29-361                                             \\
                                        & \textbf{ShakeGestureWiimoteZ}        & 100                                                  & 10                                             & 1                                                     & 40-385                                             \\
                                        & \textbf{ToeSegmentation1}            & 268                                                  & 2                                              & 1                                                     & 277                                                \\
                                        & \textbf{ToeSegmentation2}            & 166                                                  & 2                                              & 1                                                     & 343                                                \\ \midrule
\multirow{6}{*}{\textbf{ECG \& EEG}}    & \textbf{Blink}                       & 950                                                  & 2                                              & 4                                                     & 510                                                \\
                                        & \textbf{ECG200}                      & 200                                                  & 2                                              & 1                                                     & 96                                                 \\
                                        & \textbf{SelfRegulationSCP1}          & 561                                                  & 2                                              & 6                                                     & 896                                                \\
                                        & \textbf{SelfRegulationSCP2}          & 380                                                  & 2                                              & 7                                                     & 1152                                               \\
                                        & \textbf{StandWalkJump}               & 27                                                   & 3                                              & 4                                                     & 2500                                               \\
                                        & \textbf{TwoLeadECG}                  & 1162                                                 & 2                                              & 1                                                     & 82                                                 \\ \midrule
\multirow{6}{*}{\textbf{Sensor}}        & \textbf{DodgerLoopDay}               & 158                                                  & 7                                              & 1                                                     & 288                                                \\
                                        & \textbf{DodgerLoopGame}              & 158                                                  & 2                                              & 1                                                     & 288                                                \\
                                        & \textbf{DodgerLoopWeekend}           & 158                                                  & 2                                              & 1                                                     & 288                                                \\
                                        & \textbf{Lightning2}                  & 121                                                  & 2                                              & 1                                                     & 637                                                \\
                                        & \textbf{Lightning7}                  & 143                                                  & 7                                              & 1                                                     & 319                                                \\
                                        & \textbf{Trace}                       & 200                                                  & 4                                              & 1                                                     & 275                                                \\ \bottomrule
\end{tabular}
\caption{Description of datasets for the classification task}\label{tab:data}
\end{table*}

\subsection{Experimental Protocol}
We utilized the source codes for \texttt{torchcde} library\footnote{\url{https://github.com/patrick-kidger/torchcde}}~\citep{kidger2020neural,morrill2021neural},and Neural Flow\footnote{\url{https://github.com/mbilos/neural-flows-experiments}}~\citep{bilovs2021neural} for the proposed method. 
In case of benchmark, we used implementation of RNN, LSTM, and GRU using \texttt{pytorch}\footnote{\url{https://pytorch.org/}}~\citep{paszke2019pytorch}.
Also, this research incorporated the primary Neural CDE algorithms from their original repository\footnote{\url{https://github.com/patrick-kidger/NeuralCDE}}~\citep{kidger2020neural}, which included methodologies such as GRU-$\Delta t$, GRU-D, GRU-ODE, ODE-RNN, and Neural CDE. Further, this study leveraged the primary source code of MTAN\footnote{\url{https://github.com/reml-lab/mTAN}}~\citep{shukla2021multi}, MIAM\footnote{\url{https://github.com/ku-milab/MIAM}}~\citep{lee2022multi}, ODE-LSTM\footnote{\url{https://github.com/mlech26l/ode-lstms}}~\citep{lechner2020learning}, Neural RDE\footnote{\url{https://github.com/jambo6/neuralRDEs}}~\citep{morrill2021neural}, ANCDE\footnote{\url{https://github.com/sheoyon-jhin/ANCDE}}~\citep{jhin2024attentive}, EXIT\footnote{\url{https://github.com/sheoyon-jhin/EXIT}}~\citep{jhin2022exit}, and LEAP\footnote{\url{https://github.com/alflsowl12/LEAP}}~\citep{jhin2023learnable} for a comprehensive evaluation.

The integrity of the comparative analysis was upheld by utilizing the original architectural designs for all methods. Due to the variability of optimal hyperparameters across different methods and datasets, the \texttt{ray} library\footnote{\url{https://github.com/ray-project/ray}}~\citep{moritz2018ray,liaw2018tune} was employed for the refinement of these parameters. This library facilitates the automatic tuning of hyperparameters to minimize validation loss, marking a departure from previous studies that necessitated manual adjustments per dataset and model. Optimal hyperparameters determined for regular time series datasets were also applied to their irregular counterparts. The Euler(-Maruyama) method was consistently used as the ODE solver for all approaches in implicit methods.

In each model and dataset, the hyperparameters fine-tuned to minimize validation loss included the number of layers $n_l$ and the dimensions of the hidden vector $n_h$. For RNN-based methods such as RNN, LSTM, and GRU, these hyperparameters were integral to the fully-connected layer, while for NDE-based approaches, they were pertinent to the embedding layer and vector fields. 
Hyperparameter optimization was systematically conducted: the learning rate $lr$ was calibrated from $10^{-4}$ to $10^{-1}$ via log-uniform search; $n_l$ was set through grid search from the set $\left\{1,2,3,4\right\}$; and $n_h$ was determined from the set $\left\{16,32,64,128\right\}$ through grid search. The batch size was chosen from the set $\left\{16,32,64,128\right\}$, considering the total size of the data. 
All models underwent a training duration of 100 epochs, with the most effective model being chosen based on the lowest validation loss, thereby ensuring its generalizability. 

\subsection{Motivating Example: BasicMotions}
Table~\ref{tab:ex} presents the classification outcomes for both with and without missingness. Computation times are measured with regular setting using a single GPU. 
While Neural CDE generally exhibits robust performance, its extended variants tend to require increased computation time and do not consistently perform well across a broad range of tasks compared to the benchmark performance. On the other hand, Neural Flow achieves rapid convergence with reduced performance. In contrast, our proposed method not only shows superior performance but also maintains a balanced computational demand. 
\begin{table}[htbp]
\scriptsize\centering\captionsetup{skip=5pt}
\begin{tabular}{@{}llccccccc@{}}
\toprule
\multicolumn{2}{c}{\textbf{Methods}}                                                                    & \textbf{Regular} & \textbf{30\% Missing} & \textbf{50\% Missing} & \textbf{70\% Missing} & \textbf{Average} &  & \textbf{Computation time (s)} \\ \midrule
\multirow{5}{*}{\textbf{\begin{tabular}[c]{@{}l@{}}RNN\\      -based\end{tabular}}}       & RNN         & 0.750 ± 0.059    & 0.633 ± 0.095         & 0.600 ± 0.070         & 0.583 ± 0.059         & 0.642 ± 0.071    &  & 2.27 ± 0.06                   \\
                                                                                          & LSTM        & 0.833 ± 0.132    & 0.833 ± 0.156         & 0.700 ± 0.095         & 0.700 ± 0.112         & 0.767 ± 0.124    &  & 2.84 ± 0.93                   \\
                                                                                          & GRU         & 0.950 ± 0.046    & 0.967 ± 0.075         & 0.900 ± 0.091         & 0.883 ± 0.173         & 0.925 ± 0.096    &  & 2.97 ± 0.05                   \\
                                                                                          & GRU-$\Delta t$      & 0.950 ± 0.075    & 1.000 ± 0.000         & 0.967 ± 0.046         & 0.950 ± 0.046         & 0.967 ± 0.041    &  & 34.39 ± 0.27                  \\
                                                                                          & GRU-D       & 0.967 ± 0.075    & 0.950 ± 0.112         & 0.933 ± 0.070         & 0.967 ± 0.046         & 0.954 ± 0.075    &  & 42.43 ± 0.29                  \\ \midrule
\multirow{3}{*}{\textbf{\begin{tabular}[c]{@{}l@{}}Attention\\      -based\end{tabular}}} & Transformer & 0.983 ± 0.037    & 0.900 ± 0.109         & 0.967 ± 0.046         & 0.933 ± 0.091         & 0.946 ± 0.071    &  & 5.75 ± 0.39                   \\
                                                                                          & MTAN        & 0.950 ± 0.112    & 0.717 ± 0.247         & 0.717 ± 0.247         & 0.700 ± 0.046         & 0.771 ± 0.163    &  & 38.77 ± 0.71                  \\
                                                                                          & MIAM        & 0.967 ± 0.046    & 0.983 ± 0.037         & 0.933 ± 0.109         & 0.867 ± 0.126         & 0.938 ± 0.079    &  & 112.34 ± 1.43                 \\ \midrule
\multirow{3}{*}{\textbf{\begin{tabular}[c]{@{}l@{}}ODE\\      -based\end{tabular}}}       & GRU-ODE     & 1.000 ± 0.000    & 0.967 ± 0.075         & 0.983 ± 0.037         & 0.933 ± 0.070         & 0.971 ± 0.045    &  & 519.78 ± 0.34                 \\
                                                                                          & ODE-RNN     & 0.983 ± 0.037    & 1.000 ± 0.000         & 1.000 ± 0.000         & 1.000 ± 0.000         & 0.996 ± 0.009    &  & 113.29 ± 0.17                 \\
                                                                                          & ODE-LSTM    & 0.733 ± 0.109    & 0.683 ± 0.037         & 0.717 ± 0.126         & 0.567 ± 0.216         & 0.675 ± 0.122    &  & 73.03 ± 0.73                  \\ \midrule
\multirow{5}{*}{\textbf{\begin{tabular}[c]{@{}l@{}}CDE\\      -based\end{tabular}}}       & Neural CDE  & 0.983 ± 0.037    & 1.000 ± 0.000         & 0.983 ± 0.037         & 0.900 ± 0.037         & 0.967 ± 0.028    &  & 112.53 ± 0.41                 \\
                                                                                          & Neural RDE  & 0.983 ± 0.037    & 0.983 ± 0.037         & 0.950 ± 0.075         & 0.833 ± 0.083         & 0.938 ± 0.058    &  & 93.79 ± 0.44                  \\
                                                                                          & ANCDE       & 1.000 ± 0.000    & 0.967 ± 0.046         & 0.983 ± 0.037         & 0.900 ± 0.137         & 0.963 ± 0.055    &  & 360.50 ± 0.47                 \\
                                                                                          & EXIT        & 0.300 ± 0.112    & 0.450 ± 0.095         & 0.417 ± 0.102         & 0.367 ± 0.139         & 0.383 ± 0.112    &  & 334.53 ± 0.28                 \\
                                                                                          & LEAP        & 0.317 ± 0.149    & 0.300 ± 0.095         & 0.317 ± 0.124         & 0.167 ± 0.083         & 0.275 ± 0.113    &  & 163.04 ± 0.59                 \\ \midrule
\textbf{Flow-based}                                                                             & Neural Flow & 0.717 ± 0.139    & 0.433 ± 0.207         & 0.433 ± 0.109         & 0.367 ± 0.126         & 0.487 ± 0.145    &  & 4.64 ± 0.28                   \\ \midrule
\multicolumn{2}{l}{\textbf{Proposed method}}                                                            & 1.000 ± 0.000    & 1.000 ± 0.000         & 0.983 ± 0.037         & 0.983 ± 0.037         & 0.992 ± 0.019    &  & 119.08 ± 2.70                 \\ \bottomrule
\end{tabular}
\caption{Performance and computation time comparison using `BasicMotions' dataset under regular (0\% missingness) and irregular (30\%, 50\%, and 70\% missingness) scenarios. Computation time is calculated without early-stopping.}\label{tab:ex}
\end{table}

\subsection{Variations in model architecture}
Our framework's architecture allows for several potential variations, each with distinct trade-offs. In this section, we compare the versatility of our framework with two potential perspectives.
One possible modification is to utilize only $\vz(t)$ instead of $\hat{\vz}(t)$ after training the unified framework. However, this approach would significantly limit the model's expressiveness, as it fails to leverage the unique properties of the flow model, particularly its ability to model complex probability distributions through invertible transformations, as shown in Table~\ref{tab:ablation_1}.

Another consideration is the replacement of the flow model with more sophisticated architectures such as feedforward networks~\citep{hornik1989multilayer}, gated recurrent unit~\citep{chung2014empirical}, temporal convolution networks~\citep{lea2017temporal}, or transformer-based attention mechanisms~\citep{vaswani2017attention}. 
Table~\ref{tab:ablation_2} summarizes the analysis of different model components in our framework. 
While these alternatives demonstrate strong performance on regularly-sampled time series, they often struggle with irregular sampling patterns and missing data. Compared to that, the flow model in our current architecture specifically addresses these challenges. 
The flow model in our current architecture specifically addresses these challenges through its continuous and invertible nature, making it particularly well-suited for handling irregular time series data with missing observations.

\begin{table}[htbp]
\scriptsize\centering\captionsetup{skip=5pt}
\begin{tabular}{@{}lcccccccc@{}}
\toprule
\multicolumn{1}{c}{\multirow{2.5}{*}{\textbf{\textbf{Configuration}}}}                                  & \multicolumn{2}{c}{\textbf{Regular}} & \multicolumn{2}{c}{\textbf{30\% Missing}} & \multicolumn{2}{c}{\textbf{50\% Missing}} & \multicolumn{2}{c}{\textbf{70\% Missing}} \\ \cmidrule(lr){2-3} \cmidrule(lr){4-5} \cmidrule(lr){6-7} \cmidrule(lr){8-9} 
\multicolumn{1}{c}{}                                                                   & \textbf{Accuracy}   & \textbf{Loss}  & \textbf{Accuracy}     & \textbf{Loss}     & \textbf{Accuracy}     & \textbf{Loss}     & \textbf{Accuracy}     & \textbf{Loss}     \\ \midrule
\textbf{\begin{tabular}[c]{@{}l@{}}Proposed   method\\      (Neural CDE)\end{tabular}} & 1.000 (0.000)                         & 0.007 (0.008)                     & 1.000 (0.000)                         & 0.021 (0.027)                     & 0.983 (0.037)                         & 0.044 (0.063)                     & 0.983 (0.037)                         & 0.069 (0.141)                     \\ 
-- using $\vz(t)$ instead                                                              & 1.000 (0.000)                         & 0.152 (0.087)                     & 1.000 (0.000)                         & 0.144 (0.119)                     & 1.000 (0.000)                         & 0.126 (0.126)                     & 0.883 (0.095)                         & 0.346 (0.174)                     \\ \midrule
Neural CDE                                                                             & 0.983 (0.037)                         & 0.037 (0.056)                     & 1.000 (0.000)                         & 0.048 (0.043)                     & 0.983 (0.037)                         & 0.071 (0.061)                     & 0.900 (0.037)                         & 0.433 (0.242)                     \\ \midrule
Neural Flow                                                                            & 0.717 (0.139)                         & 0.969 (0.340)                     & 0.433 (0.207)                         & 1.384 (0.285)                     & 0.433 (0.109)                         & 1.382 (0.382)                     & 0.367 (0.126)                         & 1.521 (0.343)                     \\ \bottomrule
\end{tabular}
\caption{Comparative study of latent value selection using `BasicMotions' dataset}\label{tab:ablation_1}
\end{table}
\begin{table}[htbp]
\scriptsize\centering\captionsetup{skip=5pt}
\begin{tabular}{@{}lcccccccc@{}}
\toprule
\multicolumn{1}{c}{\multirow{2.5}{*}{\textbf{\textbf{Configuration}}}}                                  & \multicolumn{2}{c}{\textbf{Regular}} & \multicolumn{2}{c}{\textbf{30\% Missing}} & \multicolumn{2}{c}{\textbf{50\% Missing}} & \multicolumn{2}{c}{\textbf{70\% Missing}} \\ \cmidrule(lr){2-3} \cmidrule(lr){4-5} \cmidrule(lr){6-7} \cmidrule(lr){8-9} 
\multicolumn{1}{c}{}                                                                   & \textbf{Accuracy}   & \textbf{Loss}  & \textbf{Accuracy}     & \textbf{Loss}     & \textbf{Accuracy}     & \textbf{Loss}     & \textbf{Accuracy}     & \textbf{Loss}     \\ \midrule
\textbf{\begin{tabular}[c]{@{}l@{}}Proposed   method\\      (Neural CDE)\end{tabular}} & 1.000 (0.000)                         & 0.007 (0.008)                     & 1.000 (0.000)                         & 0.021 (0.027)                     & 0.983 (0.037)                         & 0.044 (0.063)                     & 0.983 (0.037)                         & 0.069 (0.141)                     \\ \hdashline
-- Baseline                                                                            & 0.983 (0.037)                         & 0.037 (0.056)                     & 1.000 (0.000)                         & 0.048 (0.043)                     & 0.983 (0.037)                         & 0.071 (0.061)                     & 0.900 (0.037)                         & 0.433 (0.242)                     \\
-- Conventional MLP                                                                    & 0.950 (0.075)                         & 0.131 (0.104)                     & 0.933 (0.037)                         & 0.263 (0.179)                     & 0.950 (0.046)                         & 0.167 (0.113)                     & 0.933 (0.037)                         & 0.214 (0.053)                     \\
-- ResNet Flow                                                                         & 0.983 (0.037)                         & 0.048 (0.053)                     & 0.967 (0.075)                         & 0.078 (0.121)                     & 0.917 (0.102)                         & 0.135 (0.121)                     & 1.000 (0.000)                         & 0.041 (0.035)                     \\
-- GRU Flow                                                                            & 0.967 (0.046)                         & 0.093 (0.073)                     & 0.950 (0.075)                         & 0.116 (0.107)                     & 0.967 (0.046)                         & 0.158 (0.096)                     & 0.983 (0.037)                         & 0.088 (0.057)                     \\
-- Coupling Flow                                                                       & 1.000 (0.000)                         & 0.007 (0.008)                     & 1.000 (0.000)                         & 0.021 (0.027)                     & 0.983 (0.037)                         & 0.044 (0.063)                     & 0.983 (0.037)                         & 0.069 (0.141)                     \\ \midrule
-- Feedforward network                                                                 & 1.000 (0.000)                         & 0.023 (0.014)                     & 0.983 (0.037)                         & 0.066 (0.042)                     & 0.900 (0.109)                         & 0.314 (0.278)                     & 0.850 (0.070)                         & 0.432 (0.152)                     \\
-- Gated recurrent unit                                                                & 0.983 (0.037)                         & 0.050 (0.063)                     & 0.883 (0.075)                         & 0.402 (0.339)                     & 0.883 (0.046)                         & 0.333 (0.176)                     & 0.867 (0.075)                         & 0.450 (0.199)                     \\
-- Temporal convolution                                                                & 0.850 (0.149)                         & 0.386 (0.298)                     & 0.850 (0.091)                         & 0.455 (0.383)                     & 0.900 (0.070)                         & 0.408 (0.211)                     & 0.767 (0.109)                         & 0.575 (0.272)                     \\
-- Transformer                                                                         & 0.917 (0.083)                         & 0.254 (0.252)                     & 0.950 (0.112)                         & 0.292 (0.639)                     & 0.950 (0.075)                         & 0.195 (0.286)                     & 0.867 (0.095)                         & 0.521 (0.390)                     \\ \bottomrule
\end{tabular}
\caption{Comparative study of component selection using `BasicMotions' dataset}\label{tab:ablation_2}
\end{table}
\begin{table}[htbp]
\scriptsize\centering\captionsetup{skip=5pt}
\begin{tabular}{@{}lcccccccc@{}}
\toprule
\multicolumn{1}{c}{\multirow{2.5}{*}{\textbf{\textbf{Configuration}}}}                                  & \multicolumn{2}{c}{\textbf{Regular}} & \multicolumn{2}{c}{\textbf{30\% Missing}} & \multicolumn{2}{c}{\textbf{50\% Missing}} & \multicolumn{2}{c}{\textbf{70\% Missing}} \\ \cmidrule(lr){2-3} \cmidrule(lr){4-5} \cmidrule(lr){6-7} \cmidrule(lr){8-9} 
\multicolumn{1}{c}{}                                                                   & \textbf{Accuracy}   & \textbf{Loss}  & \textbf{Accuracy}     & \textbf{Loss}     & \textbf{Accuracy}     & \textbf{Loss}     & \textbf{Accuracy}     & \textbf{Loss}     \\ \midrule
\textbf{\begin{tabular}[c]{@{}l@{}}Proposed   method\\      (Neural CDE)\end{tabular}} & 0.724 (0.090)       & 0.631 (0.145)  & 0.720 (0.088)         & 0.703 (0.180)     & 0.691 (0.091)         & 0.712 (0.173)     & 0.697 (0.098)         & 0.698 (0.176)     \\ \hdashline
-- Baseline                                                                            & 0.681 (0.073)       & 0.719 (0.252)  & 0.672 (0.068)         & 0.739 (0.253)     & 0.661 (0.070)         & 0.768 (0.248)     & 0.652 (0.091)         & 0.836 (0.303)     \\
-- Conventional MLP                                                                    & 0.592 (0.084)       & 0.812 (0.143)  & 0.578 (0.091)         & 0.870 (0.152)     & 0.578 (0.087)         & 0.888 (0.163)     & 0.582 (0.085)         & 0.879 (0.136)     \\
-- ResNet Flow                                                                         & 0.678 (0.084)       & 0.714 (0.141)  & 0.662 (0.094)         & 0.791 (0.181)     & 0.638 (0.086)         & 0.791 (0.160)     & 0.625 (0.100)         & 0.812 (0.138)     \\
-- GRU Flow                                                                            & 0.669 (0.079)       & 0.685 (0.135)  & 0.662 (0.078)         & 0.776 (0.184)     & 0.656 (0.074)         & 0.803 (0.178)     & 0.652 (0.083)         & 0.810 (0.191)     \\
-- Coupling Flow                                                                       & 0.675 (0.098)       & 0.715 (0.166)  & 0.673 (0.090)         & 0.745 (0.197)     & 0.667 (0.096)         & 0.769 (0.185)     & 0.662 (0.102)         & 0.758 (0.194)     \\ \midrule
\textbf{\begin{tabular}[c]{@{}l@{}}Proposed   method\\      (Neural ODE)\end{tabular}} & 0.540 (0.059)       & 0.948 (0.064)  & 0.534 (0.076)         & 0.971 (0.078)     & 0.535 (0.064)         & 0.977 (0.077)     & 0.531 (0.066)         & 0.978 (0.070)     \\ \hdashline
-- Baseline                                                                            & 0.511 (0.069)       & 0.973 (0.080)  & 0.503 (0.072)         & 0.983 (0.078)     & 0.489 (0.068)         & 0.985 (0.084)     & 0.506 (0.066)         & 0.982 (0.069)     \\
-- Conventional MLP                                                                    & 0.521 (0.062)       & 0.957 (0.065)  & 0.503 (0.067)         & 0.981 (0.082)     & 0.494 (0.063)         & 0.993 (0.085)     & 0.496 (0.071)         & 0.988 (0.071)     \\
-- ResNet Flow                                                                         & 0.528 (0.071)       & 0.960 (0.062)  & 0.505 (0.059)         & 0.978 (0.087)     & 0.503 (0.051)         & 0.987 (0.081)     & 0.504 (0.059)         & 0.987 (0.076)     \\
-- GRU Flow                                                                            & 0.525 (0.057)       & 0.965 (0.063)  & 0.530 (0.083)         & 0.974 (0.083)     & 0.520 (0.075)         & 0.999 (0.104)     & 0.515 (0.068)         & 0.986 (0.068)     \\
-- Coupling Flow                                                                       & 0.532 (0.061)       & 0.944 (0.064)  & 0.528 (0.077)         & 0.976 (0.075)     & 0.522 (0.057)         & 0.983 (0.072)     & 0.509 (0.062)         & 0.980 (0.068)     \\ \midrule
\textbf{\begin{tabular}[c]{@{}l@{}}Proposed   method\\      (Neural SDE)\end{tabular}} & 0.543 (0.066)       & 0.958 (0.083)  & 0.531 (0.063)         & 0.971 (0.063)     & 0.537 (0.073)         & 0.974 (0.071)     & 0.541 (0.061)         & 0.970 (0.068)     \\ \hdashline
-- Baseline                                                                            & 0.521 (0.063)       & 0.961 (0.073)  & 0.502 (0.059)         & 0.977 (0.078)     & 0.505 (0.057)         & 0.983 (0.072)     & 0.507 (0.064)         & 0.992 (0.069)     \\
-- Conventional MLP                                                                    & 0.516 (0.065)       & 0.961 (0.084)  & 0.495 (0.070)         & 0.995 (0.075)     & 0.511 (0.073)         & 0.984 (0.074)     & 0.503 (0.068)         & 0.983 (0.073)     \\
-- ResNet Flow                                                                         & 0.526 (0.066)       & 0.965 (0.064)  & 0.518 (0.070)         & 0.964 (0.075)     & 0.513 (0.061)         & 0.972 (0.079)     & 0.522 (0.060)         & 0.973 (0.073)     \\
-- GRU Flow                                                                            & 0.520 (0.058)       & 0.980 (0.086)  & 0.528 (0.069)         & 0.980 (0.075)     & 0.522 (0.078)         & 0.982 (0.063)     & 0.515 (0.066)         & 0.984 (0.070)     \\
-- Coupling Flow                                                                       & 0.528 (0.071)       & 0.961 (0.087)  & 0.513 (0.068)         & 0.976 (0.076)     & 0.511 (0.062)         & 0.983 (0.083)     & 0.526 (0.071)         & 0.981 (0.070)     \\ \bottomrule
\end{tabular}
\caption{Ablation study of the proposed method on 18 datasets under regular and three missing rates (Values in parentheses show average of 18 standard deviations)}\label{tab:ablation_all}
\end{table}

\subsection{Detailed Results of Ablation Study}\label{appendix:ablation}
Table~\ref{tab:ablation_all} presents a comprehensive ablation study of our \texttt{DualDynamics} framework, comparing various combinations of implicit components (Neural ODE, CDE, and SDE) and explicit components (ResNet Flow, GRU Flow, and Coupling Flow) against a non-invertible conventional MLP baseline. Neural CDEs consistently outperform Neural ODEs and SDEs, aligning with previous findings in the literature. While the optimal flow configuration varies across datasets, all flow models demonstrate superior performance compared to the MLP baseline. This consistent improvement underscores the efficacy of our framework's synergistic integration of implicit and explicit components.

\subsection{Detailed Results Three Domains}\label{appendix:domain}
Our investigation involved 18 datasets classified into three domains. 
We collate the outcomes across three domains: Motion \& HAR, ECG \& EEG, and Sensor datasets, in Tables~\ref{tab:result1}, \ref{tab:result2}, and \ref{tab:result3}. 
We observe variability in the performance metrics due to the unique characteristics of each domain. 
Overall, the proposed methodology demonstrates superior performance relative to benchmark models, in the different domains and missing rate scenarios.

\begin{table}[ht]
\scriptsize\centering\captionsetup{skip=5pt}
\begin{tabular}{@{}lcccccccccccccc@{}}
\toprule
\multirow{2.5}{*}{\textbf{Methods}} & \multicolumn{2}{c}{\textbf{Regular}} & \textbf{} & \multicolumn{2}{c}{\textbf{30\% Missing}} & \textbf{} & \multicolumn{2}{c}{\textbf{50\% Missing}} & \textbf{} & \multicolumn{2}{c}{\textbf{70\% Missing}} & \textbf{} & \multicolumn{2}{c}{\textbf{Average}} \\ \cmidrule(lr){2-3} \cmidrule(lr){5-6} \cmidrule(lr){8-9} \cmidrule(lr){11-12} \cmidrule(l){14-15} 
                                  & \textbf{Accuracy}   & \textbf{Rank}  & \textbf{} & \textbf{Accuracy}     & \textbf{Rank}     & \textbf{} & \textbf{Accuracy}     & \textbf{Rank}     & \textbf{} & \textbf{Accuracy}     & \textbf{Rank}     & \textbf{} & \textbf{Accuracy}   & \textbf{Rank}  \\ \midrule
RNN                               & 0.518 (0.062)       & 13.3           &           & 0.445 (0.077)         & 15.8              &           & 0.427 (0.061)         & 15.6              &           & 0.415 (0.070)         & 14.5              &           & 0.451 (0.067)       & 14.8           \\
LSTM                              & 0.568 (0.089)       & 11.6           &           & 0.522 (0.092)         & 11.9              &           & 0.494 (0.094)         & 12.3              &           & 0.467 (0.090)         & 13.3              &           & 0.513 (0.091)       & 12.3           \\
GRU                               & 0.696 (0.095)       & 7.2            &           & 0.629 (0.076)         & 10.3              &           & 0.612 (0.089)         & 9.6               &           & 0.631 (0.111)         & 7.8               &           & 0.642 (0.093)       & 8.7            \\
GRU-$\Delta t$                            & 0.613 (0.076)       & 9.3            &           & 0.624 (0.079)         & 7.6               &           & 0.663 (0.064)         & 7.3               &           & 0.624 (0.077)         & 9.3               &           & 0.631 (0.074)       & 8.4            \\
GRU-D                             & 0.587 (0.077)       & 11.6           &           & 0.563 (0.089)         & 11.3              &           & 0.540 (0.056)         & 13.2              &           & 0.564 (0.054)         & 11.2              &           & 0.563 (0.069)       & 11.8           \\
Transformer                       & 0.772 (0.060)       & 3.2            &           & 0.703 (0.070)         & 4.8               &           & 0.718 (0.079)         & 4.3               &           & 0.689 (0.069)         & 4.3               &           & 0.720 (0.070)       & 4.1            \\
MTAN                              & 0.603 (0.096)       & 12.7           &           & 0.583 (0.116)         & 9.9               &           & 0.587 (0.099)         & 9.2               &           & 0.593 (0.063)         & 9.2               &           & 0.591 (0.093)       & 10.2           \\
MIAM                              & 0.590 (0.091)       & 9.0            &           & 0.603 (0.063)         & 7.8               &           & 0.589 (0.066)         & 8.6               &           & 0.562 (0.044)         & 9.8               &           & 0.586 (0.066)       & 8.8            \\
GRU-ODE                           & 0.662 (0.081)       & 8.8            &           & 0.647 (0.091)         & 9.5               &           & 0.656 (0.072)         & 8.7               &           & 0.657 (0.084)         & 7.0               &           & 0.656 (0.082)       & 8.5            \\
ODE-RNN                           & 0.621 (0.087)       & 9.8            &           & 0.603 (0.078)         & 8.6               &           & 0.588 (0.095)         & 8.8               &           & 0.627 (0.064)         & 7.6               &           & 0.610 (0.081)       & 8.7            \\
ODE-LSTM                          & 0.550 (0.083)       & 12.8           &           & 0.507 (0.080)         & 12.8              &           & 0.475 (0.102)         & 13.7              &           & 0.426 (0.094)         & 13.2              &           & 0.489 (0.089)       & 13.1           \\
Neural CDE                        & 0.751 (0.096)       & 5.9            &           & 0.728 (0.080)         & 5.5               &           & 0.717 (0.100)         & 5.9               &           & 0.738 (0.111)         & 6.2               &           & 0.733 (0.097)       & 5.9            \\
Neural RDE                        & 0.702 (0.091)       & 6.2            &           & 0.703 (0.077)         & 6.1               &           & 0.704 (0.088)         & 6.5               &           & 0.665 (0.095)         & 7.6               &           & 0.693 (0.088)       & 6.6            \\
ANCDE                             & 0.710 (0.068)       & 5.8            &           & 0.716 (0.075)         & 5.0               &           & 0.724 (0.074)         & 4.1               &           & 0.703 (0.077)         & 5.5               &           & 0.713 (0.074)       & 5.1            \\
EXIT                              & 0.548 (0.086)       & 10.8           &           & 0.584 (0.073)         & 9.3               &           & 0.567 (0.086)         & 10.4              &           & 0.534 (0.097)         & 11.8              &           & 0.558 (0.086)       & 10.6           \\
LEAP                              & 0.397 (0.084)       & 15.7           &           & 0.374 (0.086)         & 16.0              &           & 0.381 (0.077)         & 14.1              &           & 0.335 (0.062)         & 14.8              &           & 0.372 (0.077)       & 15.1           \\
Neural Flow                       & 0.527 (0.068)       & 11.3           &           & 0.420 (0.081)         & 13.3              &           & 0.389 (0.084)         & 14.1              &           & 0.360 (0.065)         & 14.3              &           & 0.424 (0.074)       & 13.2           \\ \midrule
\textbf{Proposed method}          & 0.704 (0.082)       & 6.3            &           & 0.702 (0.074)         & 5.6               &           & 0.696 (0.088)         & 4.9               &           & 0.723 (0.123)         & 3.7               &           & 0.706 (0.092)       & 5.1            \\ \bottomrule
\end{tabular}
\caption{Average classification performance on 6 datasets in Motion \& HAR 
}\label{tab:result1}
\bigskip
\scriptsize\centering\captionsetup{skip=5pt}
\begin{tabular}{@{}lcccccccccccccc@{}}
\toprule
\multirow{2.5}{*}{\textbf{Methods}} & \multicolumn{2}{c}{\textbf{Regular}} & \textbf{} & \multicolumn{2}{c}{\textbf{30\% Missing}} & \textbf{} & \multicolumn{2}{c}{\textbf{50\% Missing}} & \textbf{} & \multicolumn{2}{c}{\textbf{70\% Missing}} & \textbf{} & \multicolumn{2}{c}{\textbf{Average}} \\ \cmidrule(lr){2-3} \cmidrule(lr){5-6} \cmidrule(lr){8-9} \cmidrule(lr){11-12} \cmidrule(l){14-15} 
                                  & \textbf{Accuracy}   & \textbf{Rank}  & \textbf{} & \textbf{Accuracy}     & \textbf{Rank}     & \textbf{} & \textbf{Accuracy}     & \textbf{Rank}     & \textbf{} & \textbf{Accuracy}     & \textbf{Rank}     & \textbf{} & \textbf{Accuracy}   & \textbf{Rank}  \\ \midrule
RNN                               & 0.604 (0.075)       & 10.1           &           & 0.532 (0.058)         & 14.2              &           & 0.545 (0.062)         & 11.8              &           & 0.506 (0.056)         & 14.2              &           & 0.547 (0.063)       & 12.6           \\
LSTM                              & 0.609 (0.054)       & 10.8           &           & 0.592 (0.066)         & 8.5               &           & 0.568 (0.057)         & 9.7               &           & 0.546 (0.048)         & 10.1              &           & 0.579 (0.056)       & 9.8            \\
GRU                               & 0.692 (0.073)       & 5.8            &           & 0.704 (0.059)         & 5.2               &           & 0.658 (0.068)         & 7.9               &           & 0.661 (0.068)         & 6.9               &           & 0.679 (0.067)       & 6.5            \\
GRU-$\Delta t$                            & 0.602 (0.047)       & 12.6           &           & 0.616 (0.052)         & 10.7              &           & 0.624 (0.062)         & 8.5               &           & 0.616 (0.069)         & 9.8               &           & 0.614 (0.057)       & 10.4           \\
GRU-D                             & 0.621 (0.056)       & 10.0           &           & 0.606 (0.066)         & 11.0              &           & 0.618 (0.055)         & 9.8               &           & 0.585 (0.061)         & 12.3              &           & 0.608 (0.060)       & 10.8           \\
Transformer                       & 0.693 (0.067)       & 8.9            &           & 0.643 (0.056)         & 10.2              &           & 0.621 (0.066)         & 10.8              &           & 0.621 (0.088)         & 11.1              &           & 0.645 (0.069)       & 10.3           \\
MTAN                              & 0.688 (0.081)       & 8.2            &           & 0.680 (0.115)         & 5.8               &           & 0.676 (0.093)         & 4.9               &           & 0.699 (0.093)         & 4.8               &           & 0.686 (0.096)       & 5.9            \\
MIAM                              & 0.634 (0.062)       & 11.3           &           & 0.651 (0.063)         & 7.9               &           & 0.644 (0.055)         & 7.2               &           & 0.637 (0.058)         & 7.8               &           & 0.642 (0.060)       & 8.5            \\
GRU-ODE                           & 0.650 (0.057)       & 8.1            &           & 0.661 (0.057)         & 6.8               &           & 0.653 (0.063)         & 7.5               &           & 0.633 (0.080)         & 8.4               &           & 0.649 (0.064)       & 7.7            \\
ODE-RNN                           & 0.664 (0.068)       & 6.3            &           & 0.650 (0.050)         & 7.4               &           & 0.644 (0.048)         & 9.0               &           & 0.647 (0.050)         & 6.2               &           & 0.651 (0.054)       & 7.2            \\
ODE-LSTM                          & 0.580 (0.058)       & 9.6            &           & 0.546 (0.065)         & 12.3              &           & 0.550 (0.054)         & 12.8              &           & 0.545 (0.036)         & 12.8              &           & 0.555 (0.053)       & 11.8           \\
Neural CDE                        & 0.643 (0.040)       & 8.1            &           & 0.632 (0.038)         & 8.3               &           & 0.625 (0.041)         & 8.8               &           & 0.596 (0.069)         & 8.3               &           & 0.624 (0.047)       & 8.3            \\
Neural RDE                        & 0.591 (0.065)       & 12.5           &           & 0.621 (0.060)         & 8.6               &           & 0.602 (0.070)         & 9.3               &           & 0.577 (0.073)         & 9.3               &           & 0.598 (0.067)       & 9.9            \\
ANCDE                             & 0.617 (0.075)       & 9.3            &           & 0.612 (0.060)         & 9.3               &           & 0.596 (0.059)         & 11.3              &           & 0.590 (0.072)         & 7.9               &           & 0.604 (0.066)       & 9.4            \\
EXIT                              & 0.599 (0.092)       & 10.8           &           & 0.552 (0.084)         & 12.8              &           & 0.563 (0.092)         & 11.2              &           & 0.574 (0.044)         & 10.2              &           & 0.572 (0.078)       & 11.2           \\
LEAP                              & 0.572 (0.051)       & 12.3           &           & 0.527 (0.080)         & 14.3              &           & 0.546 (0.093)         & 11.8              &           & 0.540 (0.086)         & 12.5              &           & 0.546 (0.077)       & 12.7           \\
Neural Flow                       & 0.565 (0.047)       & 11.7           &           & 0.547 (0.053)         & 12.4              &           & 0.522 (0.046)         & 13.3              &           & 0.516 (0.047)         & 12.4              &           & 0.538 (0.048)       & 12.5           \\ \midrule
\textbf{Proposed method}          & 0.708 (0.119)       & 4.8            &           & 0.747 (0.086)         & 5.5               &           & 0.703 (0.086)         & 5.5               &           & 0.683 (0.096)         & 6.3               &           & 0.710 (0.097)       & 5.5            \\ \bottomrule
\end{tabular}
\caption{Average classification performance on 6 datasets in ECG \& EEG 
}\label{tab:result2}
\end{table}
\begin{table}[htbp]
\scriptsize\centering\captionsetup{skip=5pt}
\begin{tabular}{@{}lcccccccccccccc@{}}
\toprule
\multirow{2.5}{*}{\textbf{Methods}} & \multicolumn{2}{c}{\textbf{Regular}} & \textbf{} & \multicolumn{2}{c}{\textbf{30\% Missing}} & \textbf{} & \multicolumn{2}{c}{\textbf{50\% Missing}} & \textbf{} & \multicolumn{2}{c}{\textbf{70\% Missing}} & \textbf{} & \multicolumn{2}{c}{\textbf{Average}} \\ \cmidrule(lr){2-3} \cmidrule(lr){5-6} \cmidrule(lr){8-9} \cmidrule(lr){11-12} \cmidrule(l){14-15} 
                                  & \textbf{Accuracy}   & \textbf{Rank}  & \textbf{} & \textbf{Accuracy}     & \textbf{Rank}     & \textbf{} & \textbf{Accuracy}     & \textbf{Rank}     & \textbf{} & \textbf{Accuracy}     & \textbf{Rank}     & \textbf{} & \textbf{Accuracy}   & \textbf{Rank}  \\ \midrule
RNN                               & 0.557 (0.078)       & 12.4           &           & 0.476 (0.089)         & 14.1              &           & 0.441 (0.123)         & 14.7              &           & 0.437 (0.079)         & 14.6              &           & 0.478 (0.092)       & 13.9           \\
LSTM                              & 0.588 (0.059)       & 10.5           &           & 0.542 (0.067)         & 10.2              &           & 0.484 (0.069)         & 12.3              &           & 0.503 (0.065)         & 10.4              &           & 0.529 (0.065)       & 10.8           \\
GRU                               & 0.633 (0.073)       & 9.0            &           & 0.585 (0.059)         & 9.7               &           & 0.563 (0.072)         & 9.5               &           & 0.526 (0.084)         & 11.0              &           & 0.577 (0.072)       & 9.8            \\
GRU-$\Delta t$                            & 0.673 (0.072)       & 7.9            &           & 0.666 (0.075)         & 6.3               &           & 0.668 (0.080)         & 5.8               &           & 0.708 (0.075)         & 4.4               &           & 0.679 (0.076)       & 6.1            \\
GRU-D                             & 0.569 (0.130)       & 11.1           &           & 0.569 (0.105)         & 9.5               &           & 0.583 (0.115)         & 8.8               &           & 0.647 (0.070)         & 7.0               &           & 0.592 (0.105)       & 9.1            \\
Transformer                       & 0.695 (0.063)       & 8.3            &           & 0.643 (0.073)         & 9.2               &           & 0.669 (0.081)         & 7.1               &           & 0.618 (0.079)         & 9.3               &           & 0.656 (0.074)       & 8.4            \\
MTAN                              & 0.671 (0.088)       & 7.3            &           & 0.639 (0.063)         & 6.4               &           & 0.629 (0.074)         & 6.3               &           & 0.635 (0.063)         & 6.6               &           & 0.643 (0.072)       & 6.6            \\
MIAM                              & 0.493 (0.085)       & 15.3           &           & 0.500 (0.132)         & 14.2              &           & 0.482 (0.093)         & 14.1              &           & 0.432 (0.086)         & 15.5              &           & 0.477 (0.099)       & 14.8           \\
GRU-ODE                           & 0.677 (0.079)       & 6.0            &           & 0.674 (0.059)         & 5.8               &           & 0.684 (0.072)         & 4.3               &           & 0.687 (0.079)         & 5.2               &           & 0.680 (0.072)       & 5.3            \\
ODE-RNN                           & 0.673 (0.098)       & 5.9            &           & 0.642 (0.101)         & 7.3               &           & 0.647 (0.115)         & 5.7               &           & 0.684 (0.062)         & 5.8               &           & 0.662 (0.094)       & 6.2            \\
ODE-LSTM                          & 0.567 (0.083)       & 11.8           &           & 0.502 (0.061)         & 13.0              &           & 0.479 (0.049)         & 13.5              &           & 0.453 (0.075)         & 14.3              &           & 0.500 (0.067)       & 13.1           \\
Neural CDE                        & 0.650 (0.082)       & 8.8            &           & 0.657 (0.087)         & 8.4               &           & 0.641 (0.069)         & 7.3               &           & 0.623 (0.094)         & 7.3               &           & 0.643 (0.083)       & 8.0            \\
Neural RDE                        & 0.653 (0.089)       & 7.0            &           & 0.620 (0.076)         & 7.7               &           & 0.594 (0.077)         & 8.7               &           & 0.580 (0.069)         & 8.7               &           & 0.612 (0.078)       & 8.0            \\
ANCDE                             & 0.660 (0.106)       & 8.8            &           & 0.656 (0.115)         & 6.8               &           & 0.598 (0.108)         & 9.3               &           & 0.601 (0.069)         & 8.6               &           & 0.629 (0.100)       & 8.4            \\
EXIT                              & 0.637 (0.082)       & 8.7            &           & 0.605 (0.107)         & 9.5               &           & 0.606 (0.081)         & 8.6               &           & 0.584 (0.074)         & 9.2               &           & 0.608 (0.086)       & 9.0            \\
LEAP                              & 0.500 (0.052)       & 15.3           &           & 0.475 (0.044)         & 14.8              &           & 0.471 (0.054)         & 14.9              &           & 0.478 (0.073)         & 13.8              &           & 0.481 (0.056)       & 14.7           \\
Neural Flow                       & 0.542 (0.063)       & 12.8           &           & 0.487 (0.067)         & 14.5              &           & 0.454 (0.043)         & 15.5              &           & 0.437 (0.049)         & 15.5              &           & 0.480 (0.055)       & 14.6           \\ \midrule
\textbf{Proposed method}          & 0.760 (0.070)       & 4.3            &           & 0.712 (0.106)         & 3.8               &           & 0.673 (0.100)         & 4.9               &           & 0.687 (0.076)         & 3.9               &           & 0.708 (0.088)       & 4.2            \\ \bottomrule
\end{tabular}
\caption{Average classification performance on 6 datasets in Sensor 
}\label{tab:result3}
\end{table}

\section{Detailed Results for ``Interpolation of Missing Data''}\label{appendix:interpolation}
The interpolation experiment was conducted using the 2012 PhysioNet Mortality dataset \cite{silva2012predicting}, which comprises multivariate time series data from ICU records. This dataset includes 37 variables collected from Intensive Care Unit (ICU) records, with each data instance capturing sporadic and infrequent measurements obtained within the initial 48-hour window post-admission to the ICU. Following the methodologies proposed by \citet{rubanova2019latent}, we adjusted the observation timestamps to the nearest minute, thereby creating up to 2880 possible measurement intervals per time series. Our experiments adhered to the methodology proposed by \citet{shukla2021multi}, involving the alteration of observation frequency from 50\% to 90\% to predict the remaining data. 

We followed experimental protocol suggested by \citet{shukla2021multi}, and its Github repository\footnote{\url{https://github.com/reml-lab/mTAN}}. For the proposed methodology, the training process spans $300$ epochs, employing a batch size of $64$ and a learning rate of $0.001$. To train our models on a dataset comprising irregularly sampled time series, we adopt a strategy from \citet{shukla2021multi}. This involves the modified VAE training method, where we optimize a normalized variational lower bound of the log marginal likelihood, grounded on the evidence lower bound (ELBO).

\begin{table}[htbp]
\scriptsize\centering\captionsetup{skip=5pt}
\begin{tabular}{@{}lccccc@{}}
\toprule
\multirow{2.5}{*}{\textbf{Configuration}}             & \multicolumn{5}{c}{\textbf{Test MSE $\times 10^{-3}$}}                                                                                   \\ \cmidrule(l){2-6} 
                                              & \textbf{50\%}          & \textbf{60\%}          & \textbf{70\%}          & \textbf{80\%}          & \textbf{90\%}          \\ \midrule
Conventional MLP                           & 3.838 ± 0.021          & 3.655 ± 0.016          & 3.508 ± 0.010          & 3.316 ± 0.062          & 3.442 ± 0.078          \\
ResNet Flow                                & \ul{3.655 ± 0.031}    & \textbf{3.537 ± 0.020} & \textbf{3.373 ± 0.042} & 3.351 ± 0.154          & \ul{3.171 ± 0.057}    \\
GRU Flow                                   & 3.754 ± 0.014          & \ul{3.562 ± 0.019}    & 3.470 ± 0.031          & \ul{3.232 ± 0.101}    & 3.222 ± 0.030          \\
Coupling Flow                              & \textbf{3.631 ± 0.049} & 3.659 ± 0.028          & \ul{3.463 ± 0.032}    & \textbf{3.224 ± 0.044} & \textbf{3.114 ± 0.050} \\ \bottomrule
\end{tabular}
\caption{Ablation study on the PhysioNet Mortality Dataset MSE $\times 10^{-3}$)}\label{tab:interpolation_ablation}
\end{table}

Table~\ref{tab:interpolation_ablation} presents an ablation study on the interpolation task. Our \texttt{DualDynamics} framework significantly improves interpolation performance, with flow model architectures consistently outperforming conventional MLPs in terms of MSE. This performance gap, observed even with comparable network capacities, underscores the importance of our framework's architectural design over mere network size in enhancing interpolation accuracy for time series data.

Hyperparameter optimization is conducted through a grid search, focusing on the number of layers $n_l \in \left\{1,2,3,4\right\}$ and hidden vector dimensions $n_h \in \left\{16,32,64,128\right\}$. 
The optimal hyperparameters for the proposed methods are remarked in bold within Table~\ref{tab:params_mse}, encompassing observed data with 50\%. We used the same hyperparameters to the remainder settings of 60\%, 70\%, 80\%, and 90\%. 
\begin{table*}[htbp]
\scriptsize\centering\captionsetup{skip=5pt}
\begin{tabular}{@{}cccccc@{}}
\toprule
\textbf{$n_l$}     & \textbf{$n_h$} & \multicolumn{1}{l}{\textbf{Conventional MLP}} & \multicolumn{1}{l}{\textbf{ResNet Flow}} & \multicolumn{1}{l}{\textbf{GRU Flow}} & \multicolumn{1}{l}{\textbf{Coupling Flow}} \\ \midrule
\multirow{4}{*}{1} & 16             & 4.574 ± 0.072                               & 4.534 ± 0.041                            & 4.753 ± 0.068                         & 4.473 ± 0.159                              \\
                   & 32             & 4.332 ± 0.121                               & 4.134 ± 0.014                            & 4.241 ± 0.058                         & 4.302 ± 0.074                              \\
                   & 64             & 3.991 ± 0.053                               & 4.370 ± 0.062                            & 3.901 ± 0.075                         & 4.276 ± 0.145                              \\
                   & 128            & 3.839 ± 0.064                               & 4.072 ± 0.042                            & 4.097 ± 0.053                         & 4.118 ± 0.083                              \\ \midrule
\multirow{4}{*}{2} & 16             & 4.593 ± 0.102                               & 4.692 ± 0.032                            & 4.685 ± 0.102                         & 4.560 ± 0.056                              \\
                   & 32             & 4.253 ± 0.046                               & 4.208 ± 0.104                            & 4.540 ± 0.053                         & 4.519 ± 0.106                              \\
                   & 64             & 4.459 ± 0.069                               & 4.006 ± 0.062                            & 4.104 ± 0.109                         & 3.983 ± 0.047                              \\
                   & 128            & 3.876 ± 0.026                               & 4.355 ± 0.043                            & 3.778 ± 0.045                         & 3.871 ± 0.037                              \\ \midrule
\multirow{4}{*}{3} & 16             & 4.580 ± 0.088                               & 4.475 ± 0.116                            & 4.866 ± 0.054                         & 4.635 ± 0.129                              \\
                   & 32             & 4.332 ± 0.010                               & 4.096 ± 0.111                            & 4.253 ± 0.028                         & 4.146 ± 0.030                              \\
                   & 64             & 3.997 ± 0.115                               & 4.035 ± 0.153                            & 4.185 ± 0.106                         & 4.048 ± 0.119                              \\
                   & 128            & \textbf{3.838 ± 0.021}                      & \textbf{3.655 ± 0.031}                   & \textbf{3.754 ± 0.014}                & \textbf{3.631 ± 0.049}                     \\ \midrule
\multirow{4}{*}{4} & 16             & 4.552 ± 0.070                               & 4.472 ± 0.077                            & 4.744 ± 0.021                         & 4.554 ± 0.067                              \\
                   & 32             & 4.312 ± 0.044                               & 4.237 ± 0.054                            & 4.696 ± 0.185                         & 4.221 ± 0.065                              \\
                   & 64             & 4.082 ± 0.036                               & 3.995 ± 0.023                            & 4.175 ± 0.052                         & 3.955 ± 0.054                              \\
                   & 128            & 3.848 ± 0.048                               & 3.950 ± 0.042                            & 3.982 ± 0.117                         & 4.328 ± 0.048                              \\ \bottomrule
\end{tabular}
\caption{Results of hyperparameter tuning for PhysioNet Mortality Dataset (MSE $\times 10^{-3}$)}\label{tab:params_mse}
\end{table*}

\section{Detailed Results for ``Classification of Irregularly-Sampled Data''}\label{appendix:classification}

We utilized the PhysioNet Sepsis dataset~\citep{reyna2019early} to investigate classification performance with irregularly-sampled time series. The dataset comprises 40,335 patient instances from ICU, characterized by 34 temporal variables such as heart rate, oxygen saturation, and body temperature. The objective is to detect the presence or absence of sepsis in each patient.

Given that the PhysioNet dataset represents irregular time series, with only 10\% of the data points timestamped for each patient, the study addresses this irregularity through two distinct approaches to time series classification as suggested by \citet{kidger2020neural}: (i) classification utilizing observation intensity (OI) and (ii) classification without utilizing observation intensity (No OI). Observation intensity serves as an indicator of the severity of a patient's condition, and when employed, each data point in the time series is supplemented with an index reflecting this intensity. Considering the dataset's skewed distribution, performance is evaluated using the Area Under the Receiver Operating Characteristic Curve (AUROC) metric. Benchmark performance is established in \citet{jhin2022exit}. 

\begin{table}[htbp]
\scriptsize\centering\captionsetup{skip=5pt}
\begin{tabular}{@{}lccccc@{}} 
\toprule
\multirow{2.5}{*}{\textbf{Configuration}} & \multicolumn{2}{c}{\textbf{Test AUROC}} & & \multicolumn{2}{c}{\textbf{Memory (MB)}} \\ \cmidrule{2-3}\cmidrule{5-6} 
\multicolumn{1}{c}{}  & \multirow{1}{*}{\textbf{OI}}  & \multirow{1}{*}{\textbf{No OI}}  & & \multirow{1}{*}{\textbf{OI}} & \multirow{1}{*}{\textbf{No OI}} \\ \midrule
Conventional MLP & {0.912 ± 0.011} & {0.861 ± 0.016} & & 284 & 135 \\ 
ResNet Flow & {0.915 ± 0.003} & \textbf{0.875 ± 0.006} & & 230 & 156 \\ 
GRU Flow & \ul{0.916 ± 0.003} & \textbf{0.875 ± 0.004} & & 355 & 178 \\ 
Coupling Flow & \textbf{0.918 ± 0.003} & \ul{0.873 ± 0.004} & & 453 & 233 \\ 
\bottomrule
\end{tabular}
\caption{Ablation study on PhysioNet Sepsis dataset}\label{tab:sepsis_ablation}
\end{table}

We followed experimental protocol suggested by \citet{kidger2020neural}, and its Github repository\footnote{\url{https://github.com/patrick-kidger/NeuralCDE}}. For the benchmark methods, we used the reported performance in \citet{jhin2022exit}. 
Table~\ref{tab:sepsis_ablation} presents a comprehensive ablation study comparing various flow configurations integrated with Neural CDEs. 
We train for $200$ epochs with a batch size of $1024$ and a learning rate of $0.001$. 
The hyperparameters are optimized by grid search in the number of layers in vector field $n_l \in \left\{1,2,3,4\right\}$ and hidden vector dimensions $n_h \in \left\{16,32,64,128\right\}$. 
The best hyperparameters are highlighted with bold font in Table~\ref{tab:params_sepsis1} and \ref{tab:params_sepsis2}. 
Here it is acknowledged that increasing $n_h$ magnify memory and complexity instead of $n_l$. However, a more complex model does not necessarily assure superior performance.

\section{Detailed Results for ``Forecasting with the Partial Observations''}\label{appendix:forecasting}
We used two datasets for the forecasting with partial observations scenario. 
For the \textbf{MuJoCo} dataset, we followed experimental protocol suggested by in \citet{jhin2023learnable} and \citet{jhin2024attentive}. We used reported performance of benchmark methods and source code from the Github repositories\footnote{\url{https://github.com/alflsowl12/LEAP}}. 
For the \textbf{Google} dataset, We followed experimental protocol and reported performance established by \citet{jhin2022exit}. Source code from its Github repository\footnote{\url{https://github.com/sheoyon-jhin/EXIT}} was utilized to conduct experiment. 
We conducted hyperparameter tuning for regular data and the same set of hyperparameters was applied to the scenarios  entailing 30\%, 50\%, and 70\% missing data.

\paragraph{MuJoCo.}
The Multi-Joint dynamics with Contact (MuJoCo) dataset, developed by \citet{todorov2012mujoco}, is based on the Hopper configuration in the DeepMind control suite~\citep{tassa2018deepmind}. This dataset comprises 10,000 simulations, each represented as a 14-dimensional time series with 100 data points sampled at regular intervals. In our experimental protocol, the initial 50 data points in each time series were used to predict the subsequent 10 data points. To introduce complexity and simulate diverse conditions, we systematically excluded a proportion of the data points—30\%, 50\%, and 70\%—from each time series, as detailed in \citet{jhin2023learnable} and \citet{jhin2024attentive}. 

\begin{table}[htbp]
\scriptsize\centering\captionsetup{skip=5pt}
\begin{tabular}{@{}lccccc@{}}
\toprule
\multirow{2.5}{*}{\textbf{Configuration}} & \multicolumn{4}{c}{\textbf{Test MSE}}                                                             & \multirow{2.5}{*}{\textbf{Memory (MB)}} \\ \cmidrule(lr){2-5}
                                  & \textbf{Regular}       & \textbf{30\% Dropped}  & \textbf{50\% Dropped}  & \textbf{70\% Dropped}  &                                  \\ \midrule
Conventional MLP                 & \ul{0.008 ± 0.002}    & \ul{0.009 ± 0.001}    & \ul{0.009 ± 0.001}    & \ul{0.009 ± 0.001}    & 138                              \\
ResNet Flow                    & \ul{0.008 ± 0.003}    & \textbf{0.008 ± 0.001} & \textbf{0.008 ± 0.001} & \textbf{0.008 ± 0.001} & 198                              \\
GRU Flow                       & \textbf{0.006 ± 0.001} & \textbf{0.008 ± 0.001} & \textbf{0.008 ± 0.001} & \textbf{0.008 ± 0.001} & 318                              \\
Coupling Flow                  & \ul{0.008 ± 0.001}    & \textbf{0.008 ± 0.001} & \textbf{0.008 ± 0.001} & \textbf{0.008 ± 0.001} & 409                              \\ \bottomrule
\end{tabular}
\caption{Forecasting performance and memory usage on MuJoCo dataset}\label{tab:mujoco_ablation}
\end{table}

Table~\ref{tab:mujoco_ablation} summarize the ablation study of the proposed method on MuJoCo dataset. 
In the proposed method, the training spanned $500$ epochs, with a batch size configured at $1024$ and a learning rate set to $0.001$. 
The process of hyperparameter optimization was conducted through grid search, concentrating on the number of layers $n_l \in \left\{1,2,3,4\right\}$, and the dimensions of the hidden vector $n_h \in \left\{16,32,64,128\right\}$. 
The hyperparameters that were identified as optimal are conspicuously presented in bold within Table~\ref{tab:params_mujoco}. 
The same set of hyperparameters was applied to the irregularly-sampled scenarios.

\paragraph{Google.}
The Google stock data~\citep{jhin2022exit} includes transaction volumes of Google, along with metrics such as high, low, open, close, and adjusted closing prices. This dataset spans from 2011 to 2021, and the objective is to utilize historical time series data from the preceding 50 days to forecast the high, low, open, close, and adjusted closing prices for the next 10-day period. We followed the experimental protocol and reported performances of benchmark methods from \citet{jhin2022exit}. Similarly to the previous experiment, we excluded partial observations in data —30\%, 50\%, and 70\%— in order to make more challenging situations.

\begin{table}[htbp]
\scriptsize\centering\captionsetup{skip=5pt}
\begin{tabular}{@{}lccccc@{}}
\toprule
\multirow{2.5}{*}{\textbf{Configuration}} & \multicolumn{4}{c}{\textbf{Test MSE}}                                                                     & \multirow{2.5}{*}{\textbf{Memory (MB)}} \\ \cmidrule(lr){2-5}
                                  & \textbf{Regular}         & \textbf{30\% Dropped}    & \textbf{50\% Dropped}    & \textbf{70\% Dropped}    &                                  \\ \midrule
Conventional MLP                 & 0.0012 ± 0.0001          & \ul{0.0011 ± 0.0001}    & \ul{0.0011 ± 0.0001}    & \ul{0.0014 ± 0.0001}    & 52                               \\
ResNet Flow                    & \ul{0.0011 ± 0.0001}    & \textbf{0.0010 ± 0.0001} & \ul{0.0011 ± 0.0001}    & \textbf{0.0013 ± 0.0001} & 62                               \\
GRU Flow                       & 0.0012 ± 0.0001          & \textbf{0.0010 ± 0.0001} & \ul{0.0011 ± 0.0001}    & \textbf{0.0013 ± 0.0001} & 81                              \\
Coupling Flow                  & \textbf{0.0010 ± 0.0001} & \textbf{0.0010 ± 0.0001} & \textbf{0.0009 ± 0.0001} & \textbf{0.0013 ± 0.0002} & 119                              \\ \bottomrule
\end{tabular}
\caption{Forecasting performance and memory usage on Google dataset}\label{tab:google_ablation}
\end{table}

Table~\ref{tab:google_ablation} summarize the ablation study of the proposed method on Google dataset. 
In the proposed method, the training spanned $200$ epochs, with a batch size configured at $64$ and a learning rate set to $0.001$. 
The process of hyperparameter optimization was conducted through grid search, concentrating on the number of layers $n_l \in \left\{1,2,3,4\right\}$, and the dimensions of the hidden vector $n_h \in \left\{16,32,64,128\right\}$. 
Through grid search, the optimal hyperparameters are highlighted in bold within Table~\ref{tab:params_google}. 
The same set of hyperparameters was applied to the irregularly-sampled scenarios.

\begin{sidewaystable*}[htbp]
\scriptsize\centering\captionsetup{skip=5pt}
\begin{tabular}{@{}cccccccccccccc@{}}
\toprule
\multirow{2.5}{*}{\textbf{$n_l$}} & \multirow{2.5}{*}{\textbf{$n_h$}} & \multicolumn{3}{c}{\textbf{Conventional MLP}}                     & \multicolumn{3}{c}{\textbf{ResNet Flow}}                        & \multicolumn{3}{c}{\textbf{GRU Flow}}                           & \multicolumn{3}{c}{\textbf{Coupling Flow}}                      \\ \cmidrule(l){3-5} \cmidrule(l){6-8} \cmidrule(l){9-11} \cmidrule(l){12-14}
                                &                                 & Test AUROC             & Memory & \multicolumn{1}{l}{\# Params} & Test AUROC             & Memory & \multicolumn{1}{l}{\# Params} & Test AUROC             & Memory & \multicolumn{1}{l}{\# Params} & Test AUROC             & Memory & \multicolumn{1}{l}{\# Params} \\ \midrule
\multirow{4}{*}{1}              & 16                              & 0.803 ± 0.018          & 225    & 23.81                         & 0.867 ± 0.008          & 225    & 23.84                         & 0.870 ± 0.014          & 225    & 24.15                         & 0.887 ± 0.005          & 225    & 24.13                         \\
                                & 32                              & 0.851 ± 0.004          & 227    & 83.46                         & 0.878 ± 0.003          & 227    & 83.52                         & 0.878 ± 0.003          & 252    & 84.64                         & 0.880 ± 0.006          & 286    & 84.61                         \\
                                & 64                              & 0.872 ± 0.004          & 242    & 313.35                        & 0.889 ± 0.001          & 270    & 313.47                        & 0.887 ± 0.005          & 355    & 317.76                        & 0.883 ± 0.004          & 424    & 317.70                        \\
                                & 128                             & 0.872 ± 0.002          & 345    & 1215.49                       & 0.889 ± 0.006          & 401    & 1215.75                       & 0.885 ± 0.004          & 571    & 1232.51                       & 0.887 ± 0.005          & 710    & 1232.39                       \\ \midrule
\multirow{4}{*}{2}              & 16                              & 0.870 ± 0.010          & 225    & 24.35                         & 0.900 ± 0.005          & 225    & 24.39                         & 0.895 ± 0.003          & 225    & 24.42                         & 0.893 ± 0.003          & 225    & 24.67                         \\
                                & 32                              & 0.894 ± 0.003          & 227    & 85.57                         & 0.901 ± 0.003          & 227    & 85.63                         & 0.898 ± 0.003          & 252    & 85.70                         & 0.903 ± 0.006          & 293    & 86.72                         \\
                                & 64                              & 0.897 ± 0.005          & 256    & 321.67                        & 0.905 ± 0.006          & 284    & 321.79                        & 0.906 ± 0.003          & 355    & 321.92                        & 0.900 ± 0.004          & 439    & 326.02                        \\
                                & 128                             & 0.881 ± 0.005          & 373    & 1248.51                       & 0.894 ± 0.002          & 430    & 1248.77                       & 0.896 ± 0.005          & 571    & 1249.03                       & 0.895 ± 0.005          & 738    & 1265.41                       \\ \midrule
\multirow{4}{*}{3}              & 16                              & 0.892 ± 0.004          & 225    & 24.90                         & 0.905 ± 0.006          & 225    & 24.93                         & 0.905 ± 0.004          & 225    & 24.69                         & 0.905 ± 0.003          & 225    & 25.22                         \\
                                & 32                              & 0.904 ± 0.005          & 227    & 87.68                         & 0.912 ± 0.003          & 227    & 87.75                         & 0.913 ± 0.004          & 252    & 86.75                         & 0.916 ± 0.002          & 300    & 88.83                         \\
                                & 64                              & 0.907 ± 0.005          & 270    & 329.99                        & 0.912 ± 0.002          & 299    & 330.11                        & 0.913 ± 0.002          & 355    & 326.08                        & \textbf{0.918 ± 0.003} & 453    & 334.34                        \\
                                & 128                             & 0.881 ± 0.002          & 401    & 1281.54                       & 0.912 ± 0.004          & 458    & 1281.79                       & 0.908 ± 0.002          & 570    & 1265.54                       & 0.903 ± 0.006          & 766    & 1298.43                       \\ \midrule
\multirow{4}{*}{4}              & 16                              & 0.877 ± 0.007          & 225    & 25.44                         & 0.906 ± 0.008          & 225    & 25.47                         & 0.907 ± 0.004          & 225    & 24.96                         & 0.911 ± 0.005          & 229    & 25.76                         \\
                                & 32                              & 0.907 ± 0.003          & 227    & 89.79                         & \textbf{0.915 ± 0.003} & 230    & 89.86                         & 0.914 ± 0.002          & 252    & 87.81                         & 0.915 ± 0.003          & 308    & 90.95                         \\
                                & 64                              & \textbf{0.912 ± 0.011} & 284    & 338.31                        & 0.910 ± 0.004          & 313    & 338.43                        & \textbf{0.916 ± 0.003} & 355    & 330.24                        & 0.911 ± 0.004          & 467    & 342.66                        \\
                                & 128                             & 0.891 ± 0.003          & 429    & 1314.56                       & 0.909 ± 0.005          & 487    & 1314.82                       & 0.912 ± 0.004          & 571    & 1282.05                       & 0.904 ± 0.005          & 795    & 1331.46                       \\ \bottomrule
\end{tabular}
\caption{Results of hyperparameter tuning for PhysioNet Sepsis Dataset with OI (Memory usage in MB, and the number of parameter in $\times 10^3$ unit.)
}\label{tab:params_sepsis1}
\bigskip
\scriptsize\centering\captionsetup{skip=5pt}
\begin{tabular}{@{}cccccccccccccc@{}}
\toprule
\multirow{2.5}{*}{\textbf{$n_l$}} & \multirow{2.5}{*}{\textbf{$n_h$}} & \multicolumn{3}{c}{\textbf{Conventional MLP}}                     & \multicolumn{3}{c}{\textbf{ResNet Flow}}                        & \multicolumn{3}{c}{\textbf{GRU Flow}}                           & \multicolumn{3}{c}{\textbf{Coupling Flow}}                      \\ \cmidrule(l){3-5} \cmidrule(l){6-8} \cmidrule(l){9-11} \cmidrule(l){12-14}
                                &                                 & Test AUROC             & Memory & \multicolumn{1}{l}{\# Params} & Test AUROC             & Memory & \multicolumn{1}{l}{\# Params} & Test AUROC             & Memory & \multicolumn{1}{l}{\# Params} & Test AUROC             & Memory & \multicolumn{1}{l}{\# Params} \\ \midrule
\multirow{4}{*}{1}              & 16                              & 0.777 ± 0.018          & 114    & 13.76                         & 0.826 ± 0.007          & 114    & 13.79                         & 0.827 ± 0.007          & 128    & 14.10                         & 0.838 ± 0.008          & 145    & 14.08                         \\
                                & 32                              & 0.815 ± 0.007          & 121    & 46.21                         & 0.838 ± 0.008          & 135    & 46.27                         & 0.845 ± 0.006          & 178    & 47.39                         & 0.841 ± 0.007          & 212    & 47.36                         \\
                                & 64                              & 0.815 ± 0.003          & 165    & 169.47                        & 0.828 ± 0.004          & 193    & 169.60                        & 0.841 ± 0.006          & 278    & 173.89                        & 0.833 ± 0.003          & 348    & 173.83                        \\
                                & 128                             & 0.790 ± 0.018          & 259    & 649.47                        & 0.834 ± 0.004          & 316    & 649.73                        & 0.825 ± 0.008          & 485    & 666.50                        & 0.837 ± 0.010          & 624    & 666.37                        \\ \midrule
\multirow{4}{*}{2}              & 16                              & 0.835 ± 0.005          & 114    & 14.31                         & 0.861 ± 0.005          & 114    & 14.34                         & 0.860 ± 0.004          & 128    & 14.37                         & 0.861 ± 0.004          & 148    & 14.63                         \\
                                & 32                              & 0.839 ± 0.008          & 128    & 48.32                         & 0.860 ± 0.004          & 142    & 48.39                         & 0.862 ± 0.006          & 178    & 48.45                         & 0.866 ± 0.001          & 219    & 49.47                         \\
                                & 64                              & 0.830 ± 0.004          & 179    & 177.79                        & 0.856 ± 0.005          & 208    & 177.92                        & 0.851 ± 0.006          & 278    & 178.05                        & 0.861 ± 0.003          & 362    & 182.15                        \\
                                & 128                             & 0.820 ± 0.007          & 287    & 682.50                        & 0.833 ± 0.008          & 344    & 682.75                        & 0.848 ± 0.013          & 486    & 683.01                        & 0.845 ± 0.008          & 653    & 699.39                        \\ \midrule
\multirow{4}{*}{3}              & 16                              & 0.824 ± 0.018          & 114    & 14.85                         & 0.860 ± 0.008          & 114    & 14.88                         & 0.865 ± 0.011          & 128    & 14.64                         & 0.867 ± 0.007          & 152    & 15.17                         \\
                                & 32                              & \textbf{0.861 ± 0.016} & 135    & 50.43                         & 0.871 ± 0.006          & 149    & 50.50                         & \textbf{0.875 ± 0.004} & 178    & 49.51                         & 0.869 ± 0.004          & 226    & 51.59                         \\
                                & 64                              & 0.844 ± 0.006          & 193    & 186.11                        & 0.859 ± 0.010          & 222    & 186.24                        & 0.869 ± 0.003          & 278    & 182.21                        & 0.865 ± 0.008          & 376    & 190.47                        \\
                                & 128                             & 0.818 ± 0.013          & 316    & 715.52                        & 0.852 ± 0.006          & 373    & 715.78                        & 0.858 ± 0.008          & 486    & 699.52                        & 0.856 ± 0.010          & 681    & 732.42                        \\ \midrule
\multirow{4}{*}{4}              & 16                              & 0.799 ± 0.070          & 114    & 15.39                         & 0.853 ± 0.010          & 117    & 15.43                         & 0.854 ± 0.006          & 128    & 14.91                         & 0.871 ± 0.009          & 155    & 15.71                         \\
                                & 32                              & 0.848 ± 0.017          & 142    & 52.55                         & \textbf{0.875 ± 0.006} & 156    & 52.61                         & 0.867 ± 0.007          & 178    & 50.56                         & \textbf{0.873 ± 0.004} & 233    & 53.70                         \\
                                & 64                              & 0.845 ± 0.015          & 208    & 194.43                        & 0.863 ± 0.006          & 236    & 194.56                        & 0.863 ± 0.006          & 278    & 186.37                        & 0.864 ± 0.005          & 390    & 198.79                        \\
                                & 128                             & 0.823 ± 0.004          & 345    & 748.55                        & 0.846 ± 0.005          & 402    & 748.80                        & 0.858 ± 0.011          & 486    & 716.03                        & 0.858 ± 0.010          & 710    & 765.44                        \\ \bottomrule
\end{tabular}
\caption{Results of hyperparameter tuning for PhysioNet Sepsis Dataset without OI (Memory usage in MB, and the number of parameter in $\times 10^3$ unit.)
}\label{tab:params_sepsis2}
\end{sidewaystable*}

\begin{sidewaystable*}[htbp]
\scriptsize\centering\captionsetup{skip=5pt}
\begin{tabular}{@{}cccccccccccccc@{}}
\toprule
\multirow{2.5}{*}{\textbf{$n_l$}} & \multirow{2.5}{*}{\textbf{$n_h$}} & \multicolumn{3}{c}{\textbf{Conventional MLP}}                     & \multicolumn{3}{c}{\textbf{ResNet Flow}}                        & \multicolumn{3}{c}{\textbf{GRU Flow}}                           & \multicolumn{3}{c}{\textbf{Coupling Flow}}                      \\ \cmidrule(l){3-5} \cmidrule(l){6-8} \cmidrule(l){9-11} \cmidrule(l){12-14}
                                &                                 & Test MSE             & Memory & \multicolumn{1}{l}{\# Params} & Test MSE             & Memory & \multicolumn{1}{l}{\# Params} & Test MSE             & Memory & \multicolumn{1}{l}{\# Params} & Test MSE             & Memory & \multicolumn{1}{l}{\# Params} \\ \midrule
\multirow{4}{*}{1}              & 16                              & 0.020 ± 0.001          & 36     & 5.10                          & 0.018 ± 0.001          & 42     & 5.13                          & 0.017 ± 0.002          & 57     & 5.44                          & 0.016 ± 0.002          & 69     & 5.42                          \\
                                & 32                              & 0.012 ± 0.001          & 49     & 18.89                         & 0.012 ± 0.003          & 62     & 18.96                         & 0.009 ± 0.001          & 92     & 20.08                         & 0.011 ± 0.000          & 118    & 20.05                         \\
                                & 64                              & 0.009 ± 0.002          & 77     & 72.59                         & 0.009 ± 0.003          & 105    & 72.72                         & 0.006 ± 0.001          & 165    & 77.01                         & 0.008 ± 0.000          & 214    & 76.94                         \\
                                & 128                             & \textbf{0.008 ± 0.001} & 138    & 284.43                        & \textbf{0.008 ± 0.002} & 198    & 284.69                        & \textbf{0.006 ± 0.000} & 318    & 301.45                        & \textbf{0.008 ± 0.002} & 409    & 301.33                        \\ \midrule
\multirow{4}{*}{2}              & 16                              & 0.022 ± 0.002          & 38     & 5.65                          & 0.020 ± 0.001          & 45     & 5.68                          & 0.017 ± 0.003          & 57     & 5.71                          & 0.019 ± 0.002          & 72     & 5.97                          \\
                                & 32                              & 0.016 ± 0.003          & 54     & 21.01                         & 0.013 ± 0.002          & 67     & 21.07                         & 0.010 ± 0.001          & 92     & 21.13                         & 0.012 ± 0.002          & 123    & 22.16                         \\
                                & 64                              & 0.012 ± 0.001          & 87     & 80.91                         & 0.010 ± 0.002          & 114    & 81.04                         & 0.010 ± 0.001          & 165    & 81.17                         & 0.011 ± 0.001          & 225    & 85.26                         \\
                                & 128                             & 0.011 ± 0.001          & 158    & 317.45                        & 0.010 ± 0.001          & 219    & 317.71                        & 0.008 ± 0.001          & 319    & 317.97                        & 0.010 ± 0.000          & 430    & 334.35                        \\ \midrule
\multirow{4}{*}{3}              & 16                              & 0.039 ± 0.021          & 41     & 6.19                          & 0.021 ± 0.005          & 47     & 6.22                          & 0.016 ± 0.003          & 57     & 5.98                          & 0.016 ± 0.004          & 74     & 6.51                          \\
                                & 32                              & 0.033 ± 0.026          & 59     & 23.12                         & 0.014 ± 0.002          & 72     & 23.18                         & 0.011 ± 0.002          & 92     & 22.19                         & 0.012 ± 0.001          & 128    & 24.27                         \\
                                & 64                              & 0.032 ± 0.027          & 97     & 89.23                         & 0.011 ± 0.000          & 125    & 89.36                         & 0.010 ± 0.002          & 165    & 85.33                         & 0.010 ± 0.001          & 234    & 93.58                         \\
                                & 128                             & 0.013 ± 0.004          & 179    & 350.48                        & 0.011 ± 0.002          & 239    & 350.73                        & 0.008 ± 0.000          & 319    & 334.48                        & 0.011 ± 0.001          & 450    & 367.37                        \\ \midrule
\multirow{4}{*}{4}              & 16                              & 0.042 ± 0.018          & 43     & 6.73                          & 0.024 ± 0.004          & 50     & 6.77                          & 0.022 ± 0.009          & 57     & 6.25                          & 0.022 ± 0.006          & 77     & 7.05                          \\
                                & 32                              & 0.034 ± 0.026          & 64     & 25.23                         & 0.014 ± 0.001          & 78     & 25.29                         & 0.014 ± 0.003          & 92     & 23.25                         & 0.012 ± 0.003          & 132    & 26.38                         \\
                                & 64                              & 0.032 ± 0.027          & 107    & 97.55                         & 0.012 ± 0.002          & 134    & 97.68                         & 0.011 ± 0.002          & 165    & 89.49                         & 0.009 ± 0.002          & 245    & 101.90                        \\
                                & 128                             & 0.031 ± 0.028          & 199    & 383.50                        & 0.011 ± 0.002          & 260    & 383.76                        & 0.008 ± 0.001          & 319    & 350.99                        & 0.010 ± 0.000          & 471    & 400.40                        \\ \bottomrule
\end{tabular}
\caption{Results of hyperparameter tuning for MuJoCo Dataset (Memory usage in MB, and the number of parameter in $\times 10^3$ unit.)
}\label{tab:params_mujoco}
\bigskip
\scriptsize\centering\captionsetup{skip=5pt}
\begin{tabular}{@{}cccccccccccccc@{}}
\toprule
\multirow{2.5}{*}{\textbf{$n_l$}} & \multirow{2.5}{*}{\textbf{$n_h$}} & \multicolumn{3}{c}{\textbf{Conventional MLP}}                     & \multicolumn{3}{c}{\textbf{ResNet Flow}}                        & \multicolumn{3}{c}{\textbf{GRU Flow}}                           & \multicolumn{3}{c}{\textbf{Coupling Flow}}                      \\ \cmidrule(l){3-5} \cmidrule(l){6-8} \cmidrule(l){9-11} \cmidrule(l){12-14}
                                &                                 & Test MSE               & Memory & \multicolumn{1}{l}{\# Params} & Test MSE               & Memory & \multicolumn{1}{l}{\# Params} & Test MSE               & Memory & \multicolumn{1}{l}{\# Params} & Test MSE               & Memory & \multicolumn{1}{l}{\# Params} \\ \midrule
\multirow{4}{*}{1}              & 16                              & 0.0017 ± 0.0001          & 6      & 2.97                          & 0.0019 ± 0.0005          & 8      & 3.00                          & 0.0017 ± 0.0004          & 12     & 3.30                          & 0.0012 ± 0.0000          & 15     & 3.29                          \\
                                & 32                              & 0.0014 ± 0.0000          & 10     & 11.05                         & 0.0013 ± 0.0001          & 13     & 11.11                         & 0.0014 ± 0.0001          & 21     & 12.23                         & 0.0011 ± 0.0000          & 26     & 12.20                         \\
                                & 64                              & 0.0013 ± 0.0000          & 17     & 42.57                         & 0.0012 ± 0.0000          & 24     & 42.70                         & 0.0012 ± 0.0000          & 40     & 46.98                         & 0.0011 ± 0.0001          & 51     & 46.92                         \\
                                & 128                             & 0.0013 ± 0.0000          & 35     & 167.05                        & 0.0012 ± 0.0001          & 51     & 167.30                        & \textbf{0.0012 ± 0.0000} & 81     & 184.07                        & 0.0011 ± 0.0001          & 102    & 183.94                        \\ \midrule
\multirow{4}{*}{2}              & 16                              & 0.0015 ± 0.0001          & 7      & 3.51                          & 0.0022 ± 0.0009          & 8      & 3.54                          & 0.0017 ± 0.0004          & 12     & 3.58                          & 0.0011 ± 0.0001          & 15     & 3.83                          \\
                                & 32                              & 0.0014 ± 0.0000          & 11     & 13.16                         & 0.0014 ± 0.0003          & 14     & 13.22                         & 0.0014 ± 0.0001          & 21     & 13.29                         & 0.0011 ± 0.0000          & 28     & 14.31                         \\
                                & 64                              & 0.0013 ± 0.0000          & 19     & 50.89                         & 0.0012 ± 0.0001          & 26     & 51.02                         & 0.0012 ± 0.0001          & 40     & 51.14                         & 0.0011 ± 0.0000          & 54     & 55.24                         \\
                                & 128                             & 0.0013 ± 0.0000          & 41     & 200.07                        & 0.0011 ± 0.0000          & 56     & 200.33                        & 0.0012 ± 0.0000          & 81     & 200.58                        & 0.0011 ± 0.0000          & 108    & 216.97                        \\ \midrule
\multirow{4}{*}{3}              & 16                              & 0.0018 ± 0.0003          & 7      & 4.06                          & 0.0017 ± 0.0007          & 9      & 4.09                          & 0.0015 ± 0.0004          & 12     & 3.85                          & 0.0011 ± 0.0001          & 16     & 4.38                          \\
                                & 32                              & 0.0014 ± 0.0001          & 12     & 15.27                         & 0.0013 ± 0.0003          & 16     & 15.34                         & 0.0014 ± 0.0002          & 21     & 14.34                         & 0.0011 ± 0.0000          & 29     & 16.42                         \\
                                & 64                              & 0.0013 ± 0.0000          & 22     & 59.21                         & 0.0013 ± 0.0001          & 29     & 59.34                         & 0.0012 ± 0.0001          & 40     & 55.30                         & 0.0011 ± 0.0000          & 57     & 63.56                         \\
                                & 128                             & 0.0012 ± 0.0000          & 47     & 233.10                        & \textbf{0.0011 ± 0.0001} & 62     & 233.35                        & 0.0012 ± 0.0000          & 82     & 217.10                        & 0.0011 ± 0.0001          & 113    & 249.99                        \\ \midrule
\multirow{4}{*}{4}              & 16                              & 0.0015 ± 0.0001          & 8      & 4.60                          & 0.0022 ± 0.0013          & 10     & 4.63                          & 0.0018 ± 0.0006          & 12     & 4.12                          & 0.0011 ± 0.0001          & 16     & 4.92                          \\
                                & 32                              & 0.0013 ± 0.0000          & 13     & 17.38                         & 0.0016 ± 0.0005          & 17     & 17.45                         & 0.0012 ± 0.0001          & 21     & 15.40                         & 0.0011 ± 0.0000          & 30     & 18.54                         \\
                                & 64                              & 0.0013 ± 0.0000          & 25     & 67.53                         & 0.0012 ± 0.0001          & 32     & 67.66                         & 0.0012 ± 0.0001          & 40     & 59.46                         & 0.0011 ± 0.0000          & 59     & 71.88                         \\
                                & 128                             & \textbf{0.0012 ± 0.0000} & 52     & 266.12                        & 0.0011 ± 0.0000          & 68     & 266.38                        & 0.0012 ± 0.0001          & 82     & 233.61                        & \textbf{0.0010 ± 0.0000} & 119    & 283.02                        \\ \bottomrule
\end{tabular}
\caption{Results of hyperparameter tuning for Google Dataset (Memory usage in MB, and the number of parameter in $\times 10^3$ unit.)
}\label{tab:params_google}
\end{sidewaystable*}

\end{document}